\documentclass[]{spie}  

 \usepackage{multirow}
\usepackage{amsmath,amsfonts,amssymb}
\usepackage{graphicx}
\usepackage{subcaption}
\usepackage{svg}
\usepackage{physics}
\usepackage{tikz}
\usetikzlibrary{shapes.geometric}
\usetikzlibrary{positioning} 
\usepackage[colorlinks=true, allcolors=blue]{hyperref}
\usepackage{xcolor}

\DeclareMathOperator*{\argmax}{arg\,max}

\DeclareMathOperator{\vx}{\vb{x}}
\DeclareMathOperator{\vy}{\vb{y}}
\DeclareMathOperator{\vz}{\vb{z}}

\title{The Comparability of Model Fusion to Measured Data in Confuser Rejection}

\author[a]{Conor M. Flynn}
\author[b]{Christopher Ebersole}
\author[b]{Edmund Zelnio}
\affil[a]{Rensselaer Polytechnic Institute, Troy, New York, USA}
\affil[b]{Air Force Research Laboratory, WPAFB, Ohio, USA}

\authorinfo{Further author information: (Send correspondence to Christopher Ebersole)\\Christopher Ebersole: E-mail: christopher.ebersole.1@us.af.mil}

\begin{document} 
\maketitle

\begin{abstract}
Data collection has always been a major issue in the modeling and training of large deep learning networks, as no dataset can account for every slight deviation we might see in live usage. Collecting samples can be especially costly for Synthetic Aperture Radar (SAR), limiting the amount of unique targets and operating conditions we are able to observe from. To counter this lack of data, simulators have been developed utilizing the shooting and bouncing ray method to allow for the generation of synthetic SAR data on 3D models. While effective, the synthetically generated data does not perfectly correlate to the measured data leading to issues when training models solely on synthetic data. We aim to use computational power as a substitution for this lack of quality measured data, by ensembling many models trained on synthetic data. Synthetic data is also not complete, as we do not know what targets might be present in a live environment. Therefore we need to have our ensembling techniques account for these unknown targets by applying confuser rejection in which our models will reject unknown targets it is presented with, and only classify those it has been trained on.
\end{abstract}

\begin{keywords}
    Synthetic aperture radar, ensembling, confuser rejection, open set recognition, novelty detection, fusion
\end{keywords}

\section{Introduction}

High-quality data is essential for training effective deep learning models, as the performance and generalization ability of these models heavily depends on the quality, diversity, and volume of the data they are exposed to during training. However, obtaining such data can be a significant challenge especially in areas such as Synthetic Aperture Radar (SAR) \cite{bansal2022systematic}. The diversity of targets, observation angles, and other operating conditions required to achieve adequate generalization all exponentially increase the cost of acquisition of real (measured) data, making it difficult to train an accurate and robust model on the limited information collected. To counter the lack of measured SAR data available, we can utilize simulators using the shooting and bouncing ray (SBR) \cite{bhalla19983d} method to allow for the generation of synthetic SAR data on small 3D models. While this does allow for the significantly cheaper and easier acquisition of target data, the data does not come from the same distribution as measured data taken in a live environment. Therefore, we need to create techniques which use this synthetically generated data as a supplement to our limited measured data, resulting in a highly accurate and robust model.

While we can generate large quantities of synthetic data, we cannot account for all types of possible targets in a live scenario. Thus, we need to create techniques to differentiate between targets that are present in the training data and those that are not (confusers). To help determine which targets are trained on and which are confusers, we can split our dataset into three types of classes: mission targets, in-library confusers, and out-of-library confusers. Mission targets are the targets we want our system to recognize in a live setting, confidently predicting the target's class. In-library confusers are targets that we still train on, however, we present these images to our models as confusers with an obfuscated label. This is useful in confuser rejection as it trains the models to reject classes with unknown labels, giving them low confidence when making a prediction. Finally, out-of-library confusers are used in testing to ensure that our models are confidently rejecting confusers they have never seen before, reassuring that our confuser rejection training techniques were successful.

To address the issues of measured and synthetic data misalignment and confuser rejection, we use model ensembling \cite{sollich1995learning}. Model ensembling trains multiple deep learning models on the same dataset independently and then combines their individual outputs to form a joint consensus. Such a system is valuable as it generates a diverse set of model weight initialization points on the loss landscape, finding different local optima that yield different outputs \cite{fort2020deep, lakshminarayanan2017simple}. To review the effectiveness of ensembling in these problems, we implement common ensembling techniques and discuss their performance over singular model systems.

\section{Related Works}

In this section we review previous work related to these problems as well as key ideas present in this paper.

\noindent\textbf{SAR Confuser Rejection:} Casasent and Nehemiah's work discusses the implications of using the extended maximum average correlation height distortion invariant filter to determine its ability to reject confusers in the Moving and Stationary Target Acquisition and Recognition (MSTAR) public database \cite{casasent2006confuser, keydel1996mstar}. Chakravarthy, Ashby, and Zelnio discuss the usage of Adversarial Reciprocal Points Learning, which incorporates convolutional neural networks to minimize classification risk of out-of-sample targets \cite{chakravarthy2024calibrated}. Hill uses Gaussian Mixture Models on the Synthetic and Measured Paired Labeled Experiment (SAMPLE+) dataset \cite{lewis2019sar} to help predict out-of-distribution data \cite{hill2024out}. In our review we also use the SAMPLE+ dataset for testing the efficacy of our confuser rejection techniques. 

\noindent\textbf{Model Weight Variation:} Lakshminarayanan et al. and Fort et al. discuss the benefits of training various models and ensembling them \cite{lakshminarayanan2017simple, fort2020deep}. By using ensembling techniques that combine internal model sampling such as Dropout \cite{srivastava2014dropout}, as well as external model sampling referred to as Deep Ensembling by Lakshminarayanan \cite{lakshminarayanan2017simple} and as discussed in Section~\ref{sec:ensembling}, we are able to derive a robust model structure.

\noindent\textbf{Objectosphere Loss:} Finally, Dhamija et al. developed the Objectosphere Loss, which is a loss function designed to push confusers towards the center of the feature space while encouraging mission targets to have large feature magnitudes, improving feature separation between mission targets and confusers \cite{dhamija2018reducing}. This reduces confident misclassification of confusers and helps to emphasize mission target classification. We make use of this loss function and further explore it in Section~\ref{subsec:lf}.

\section{Methods}

In this section, we discuss the primary methods used in this paper to improve upon confuser rejection and mission target classification.

\subsection{Confuser Rejection}

One of the primary draw downs of deep learning networks stems from their overconfidence when classifying previously unseen targets \cite{wang2021rethinking}. Such overconfidence makes it hard to determine whether a presented target has been trained on or not, as outputs for both mission targets and confusers look indistinguishable. Therefore, we need to derive methods for differentiating between mission targets and confusers. To create a robust model architecture for isolating these types of targets we use Deep Ensembling, which combines the outputs of multiple models \cite{lakshminarayanan2017simple} as shown in Section~\ref{sec:ensembling}. Before discussing the ensembling techniques used, we first need to define a decision statistic which will be thresholded to distinguish between mission targets and confusers.

\subsubsection{Feature Magnitude}
\label{ssubsec:fm}
Our decision statistic for confuser rejection is the feature magnitude $F$, in which we observe the $L^2$ norm of the penultimate layer $P_m$. The penultimate layer refers to the input of the last fully-connected layer of each model $m$ used to produce the model's output \cite{seo2021distribution}. We observe this calculation as follows:
\begin{align}
    F_m=&||P_m||_2\\
    ||P_m||_2=&\sqrt{\sum^J_{j=0} (P_{m,j})^2}\\
    F=&\frac{1}{M}\sum_{m=0}^{M-1}F_m
\end{align}
where $F_m$ is the feature magnitude of each individual model, $M$ is the total number of models ensembled, $P_m$ is the penultimate layer of model $m$, $J$ is the size of the penultimate layer, and $P_{m,j}$ is the penultimate weight for model $m$ at spot $j$. We can use this metric to help determine how strong the classifications are for mission targets and confusers and use it as another differentiator should there be a large discrepancy in feature magnitude when making these predictions.

\subsubsection{Loss Function}
\label{subsec:lf}

To train an individual confuser rejection model we use of the Objectosphere Loss function \cite{dhamija2018reducing}. This loss function creates a loss space such that all confusers are pushed towards the center of the space and all mission targets have their feature magnitudes amplified to improve confidence. This loss is defined as:
\begin{equation}
    J_R=J_E+\lambda\begin{cases}
        \max(\xi-||F_m(\vx_n)||,0)^2 &\text{if }\vx_n\in \mathcal{C}_{mt}\\
        ||F_m(\vx_n)||^2&\text{if }\vx_n\in \mathcal{C}_{ic}\cup \mathcal{C}_{oc}
    \end{cases}
\end{equation}
\noindent where $\xi$ sets the margin or magnitude of the loss function, and $||F_m(\vx_n)||^2$ is the feature magnitude of the penultimate layer as described in Section~\ref{ssubsec:fm}, and $J_E$ is the Entropic Open-Set Loss.

By setting $\xi$ to a higher value, we encourage the model to learn a larger separation between mission targets and in-library confusers. While a positive effect, setting it too high can ``implicitly increase scaling and can impact learning rate." \cite{dhamija2018reducing} We determined that $\xi=50$ was an effective margin for our training. $F_m(\vx_n)$ refers to feature magnitude of model $m$ for the given input $\vx_n$. In our model declaration, we found 128 to be an effective number of feature dimensions for this layer. This means that our final linear layer contains weight $W$ and bias $b$ matrices of the following dimensions: $W\in\mathbb{R}^{128\times N};\;b\in\mathbb{R}^{N}$. Finally, the Entropic Open-Set Loss is defined as:
\begin{equation}
    J_E(x)=\begin{cases}
        -\log \sigma_c(\vx_n)&\text{if }\vx_n\in \mathcal{C}_{mt}\\
        -\frac{1}{C}\sum_{c=1}^C\log\sigma_c(\vx_n)&\text{if }\vx_n\in \mathcal{C}_{ic}\cup \mathcal{C}_{oc}
    \end{cases}
\end{equation}
\noindent where $\sigma_c(x)$ is the softmax activation function \cite{ganaie2022ensemble} with respect to class $c$. Note that for mission targets this loss function is minimized when the target is correctly classified, while for confusers this loss is minimized when all logits are equal.

\subsection{Ensembling}
\label{sec:ensembling}

To approach the issue of recognizing confusers and training on limited measured data, we use deep ensembling \cite{lakshminarayanan2017simple}. Ensembling works by combining the outputs of independently trained models and then creating a prediction from their joint output. By training models with randomized weight initializations, we ensure that each model will converge to a different local minima with the goal of generating diverse output predictions in the feature space \cite{fort2020deep}. We first provide a formal definition for our ensembling problem, followed by two ensembling techniques: Unweighted Model Averaging and Weighted Model Calibration \cite{ganaie2022ensemble}.

\subsubsection{Overview}
\label{subsubsec:ensm_overview}

First we formally define our ensembling problem. Let our dataset $\mathcal{D}$ consist of $N$ data points $\mathcal{D}=\left\lbrace \vx_n, y_n\right\rbrace^N_{n=1}$, where $\vx_n\in\mathbb{R}^D$ represents an image containing $D$ pixel values and $y_n$ represents the label of the associated target. Furthermore, $\mathcal{D}$ has $\mathcal{C}=\left\lbrace 0, 1, ..., C-1\right\rbrace$ labels, where $y_n\in \mathcal{C}, \forall n\in N$. Note that although targets labels are a part of the set $\mathcal{C}$, preprocessing of the data described in Section~\ref{subsec:df} turns some labels into $-1$, meaning after preprocessing $\mathcal{C}=\{0, 1, ..., C-1\}\cup \{-1\}$.

Next we give formal definitions for our model outputs. Let $\mathcal{S}=\left\lbrace \mathcal{M}_0, \mathcal{M}_1, ..., \mathcal{M}_M\right\rbrace$ be a set of $M$ models, where $\mathcal{M}_m(\vx_n)=\vy_{m,n}$ and $\vy_{m,n}\in\mathbb{R}^C$. The output, $\vy_{m,n}$ represents the traditional model output of an vector of size $C$ which containing weights associated with the predictive value of $y_n=c,\forall c\in \mathcal{C}$. We can then stack these given outputs such that $\mathcal{S}(\vx_n)=\vy_n$ where $\vy_n\in \mathbb{R}^{M\times C}$, representing a matrix containing all model outputs, each having their independent prediction values. This matrix can also be derived as $\vy_n=\left[ \vy_{0,n}, \vy_{1, n}, ..., \vy_{M,n}\right]^T$.

Finally, we derive an ensembling transformation function $\mathcal{F}:\mathbb{R}^{M\times C}\rightarrow\mathbb{R}^C$, which takes the output matrix of all the models and converts it to a single output vector of size $C$. This function, used as $\mathcal{F}(\vy_{n})=\vz_n$, outputs $\vz_n\in\mathbb{R}^C$ which is used in the label prediction function as shown in Equation~\eqref{eq:yhatpred}. The goal of $\mathcal{F}(\vy_n)$ is to reinforce the model outputs that correctly classify mission targets and reject those that either incorrectly classify mission targets or confidently classify confusers. 

Figure~\ref{fig:testflow} presents a high level overview of the provided formal derivation, depicting how an input image is classified through our ensembling system.

\tikzstyle{square} = [rectangle, minimum width=2cm, minimum height=1cm, text centered, draw=black]
\tikzstyle{arrow} = [thick,->,>=stealth]
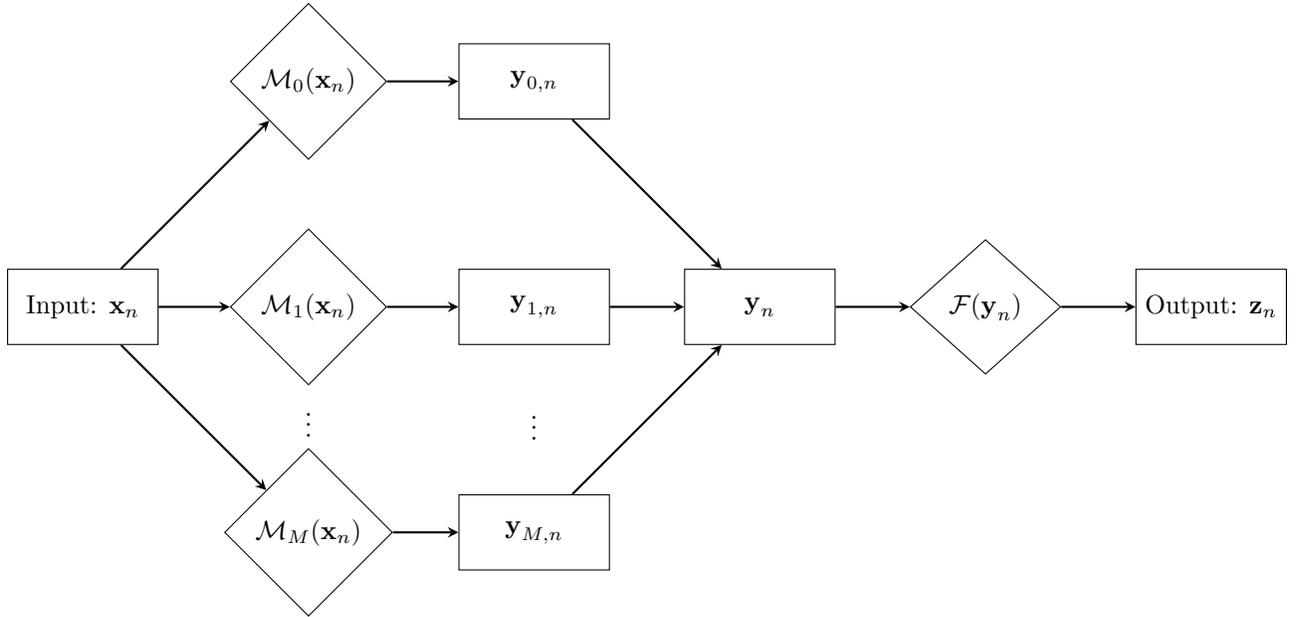
\begin{figure}[hbt!]
    \centering
    \begin{tikzpicture}
     \tikzset{node distance=3cm}
    \node (input) [square] {Input: $\vx_n$};
    \node (model1) [draw, diamond, minimum width=2cm, minimum height=1.5cm, right of=input] {$\mathcal{M}_1(\vx_n)$};
    \node (model0) [draw, diamond, minimum width=2cm, minimum height=1.5cm, above of=model1] {$\mathcal{M}_0(\vx_n)$};
    \node (modelM) [draw, diamond, minimum width=2cm, minimum height=1.5cm, below of=model1] {$\mathcal{M}_M(\vx_n)$};
    \draw [arrow] (input) -- (model0);
    \draw [arrow] (input) -- (model1);
    \draw [arrow] (input) -- (modelM);
    \path (model1) -- node[auto=false]{\vdots} (modelM);
    \node (y0) [square, right of=model0] {$\vy_{0,n}$};
    \node (y1) [square, right of=model1] {$\vy_{1,n}$};
    \node (yM) [square, right of=modelM] {$\vy_{M,n}$};
    \draw [arrow] (model0) -- (y0);
    \draw [arrow] (model1) -- (y1);
    \draw [arrow] (modelM) -- (yM);
    \path (y1) -- node[auto=false]{\vdots} (yM);
    \node (yn) [square, right of=y1] {$\vy_{n}$};
    \draw [arrow] (y0) -- (yn);
    \draw [arrow] (y1) -- (yn);
    \draw [arrow] (yM) -- (yn);
    \node (f) [draw, diamond, minimum width=2cm, minimum height=1.5cm, right of=yn] {$\mathcal{F}(\vy_n)$};
    \draw [arrow] (yn) -- (f);
    \node (z) [square, right of=f] {Output: $\vz_n$};
    \draw [arrow] (f) -- (z);
    \end{tikzpicture}
    \caption{Ensembling Flow}
    \label{fig:testflow}
\end{figure}

We deem the successful classification of a target $\left\lbrace \vx_n, y_n\right\rbrace$ if $y_n=\hat{y}_n$ which is defined as follows:
\begin{equation}
    \hat{y}_n=\begin{cases}
        -1 & \text{ if } F_n < \delta\\
        \argmax_c \vz_{n,c}&\text{ else}
    \end{cases},
    \label{eq:yhatpred}
\end{equation}
where $\vz_{n,c}$ is the logit associated with the ensembled output probability for class $c\in C$, $F_n$ is the feature magnitude for target $\vx_n$, and $\delta$ is the optimal threshold as derived by the AUROC metric in Section~\ref{subsubsec:auroc}. This provides us with $\hat{y}_n$ which is our predicted label for our image $\vx_n$.

\subsubsection{Unweighted Model Averaging}
\label{subsec:uma}

Unweighted Model Averaging works by averaging all the outputs of the models and then softmaxing them to generate a singular output \cite{ganaie2022ensemble}. We can define this technique as follows:

\begin{align}
    \mathcal{F}(\vy_{n})=&\sigma\left(\frac{1}{M}\sum_{m=0}^{M-1} \vy_{m,n}\right)
\end{align}

\noindent where $\sigma$ is the softmax activation function \cite{ganaie2022ensemble}.

While effective at determining a majority consensus on a classification, it does have a couple drawbacks. First is that several strong misclassifications from models can easily skew the results if the output is a confuser. This is especially prevalent in confuser rejection, where we want all logits to be a small number when presented with a confuser. It also does not account for the ``specialization" of given models, or their ability to classify a certain target correctly. This means a model that classifies a target correctly 20\% of the time and another model that classifies the same target correctly 80\% of the time are weighted equally in the output classification. However, ideally we would want the model that has an 80\% successful classification probability to have a higher weighting than the 20\% accuracy model to improve our chances of correctly identifying the target.

\subsubsection{Weighted Model Calibration}
\label{subsec:wmc}

To better leverage our confidence in the individual models that make up our ensemble, we derive a technique called Weighted Model Calibration. This technique trains a separate neural network on the outputs of all the models to help eliminate confident misclassifications. Note that this layer is trained independently of the models, and does not effect the weights of the individual models.

From this we create three internal fully connected layers of sizes $256, 128, \text{ and }32$ with associated weight and bias matrices $W^{(1)}\in\mathbb{R}^{(M*C)\times 256}, b^{(1)}\in\mathbb{R}^{256}, W^{(2)}\in\mathbb{R}^{256\times128}, b^{(2)}\in\mathbb{R}^{128}, W^{(3)}\in\mathbb{R}^{32\times C}, b^{(3)}\in\mathbb{R}^{C}$. We can then derive the full calculation for this Weighted Model Calibration technique as follows:
\begin{align}
    \text{flatten}(\vy_n)=& \left[\vy_{0, 0, n}, \vy_{0, 1, n},...,\vy_{0, C-1, n}, \vy_{1, 0, n}, ..., \vy_{M, C-2, n}, \vy_{M, C-1, n}\right]\\
    \text{flatten}(Y)\in&\mathbb{R}^{(M*C)}\\
    H^{(1)} =& (W^{(1)})^T (\text{flatten}(\vy_n)) + b^{(1)}\\
    H^{(2)} =& (W^{(2)})^T H^{(1)} + b^{(2)}\\
    H^{(3)} =& (W^{(3)})^T H^{(2)} + b^{(3)}\\
    \mathcal{F}(\vy_n)=&\sigma\left(H^{(3)}\right)
\end{align}

This technique has the goal of deriving more features from the image given the outputs of the various models. By introducing a second trained lightweight neural network, we can learn to dampen models that tend to have many confident misclassifications and emphasize those who have higher accuracy. While we can see this technique's efficacy in Section~\ref{ssubsec:results_wmc}, it is limited by the original issue of the paper being the lack of training data. This means that by implementing this technique we have a higher chance of over-fitting on the limited training data, as we are now introducing another trained layer that has limited weights available.


\section{Experiments}
\label{sec:experiment}

This section covers the formation of our experiments, including data preprocessing, model training specifications, and metrics of success. We then review the results of our experiments and the success of our methods using our defined metrics.

\subsection{Dataset Formation}
\label{subsec:df}

In the SAMPLE+ dataset, which we created by augmenting the standard SAMPLE dataset with MSTAR dataset targets \cite{keydel1996mstar}, we are given two types of data: synthetic and measured \cite{lewis2019sar}. Let $\mathcal{D}_s$ represent all synthetic data and $\mathcal{D}_m$ represent all measured data where: $\mathcal{D}_s\subset \mathcal{D}$, $\mathcal{D}_m\subset\mathcal{D}$, and $\mathcal{D}_s\cap \mathcal{D}_m=\varnothing$.

First we split up the classes into three categories: mission targets, in-library confusers, and out-of-library confusers. Mission targets are targets whose labels are known and trained on. In-library confusers are targets that are also trained on whose labels are obfuscated to be $-1$. Finally, out-of-library confusers are confusers whose labels are obfuscated and not trained on (only tested on). To do this, we take subsets of $\mathcal{C}$ and assign them into each of the three categories. We want more mission targets than confusers and therefore select most of the targets as mission targets. We then divide the remainder of the targets between the in-library and out-of-library confusers. For our example, we define the mission targets as images with associated labels $\mathcal{C}_{mt}=\{0, 1, ..., 9\}$, the in-library confusers as $\mathcal{C}_{ic}=\{10, ..., 12\}$ and the out-of-library confusers as $\mathcal{C}_{oc}=\{13,...,C-1\}$.

\begin{figure}[hbt!]
    \centering
    \includegraphics[scale=0.75]{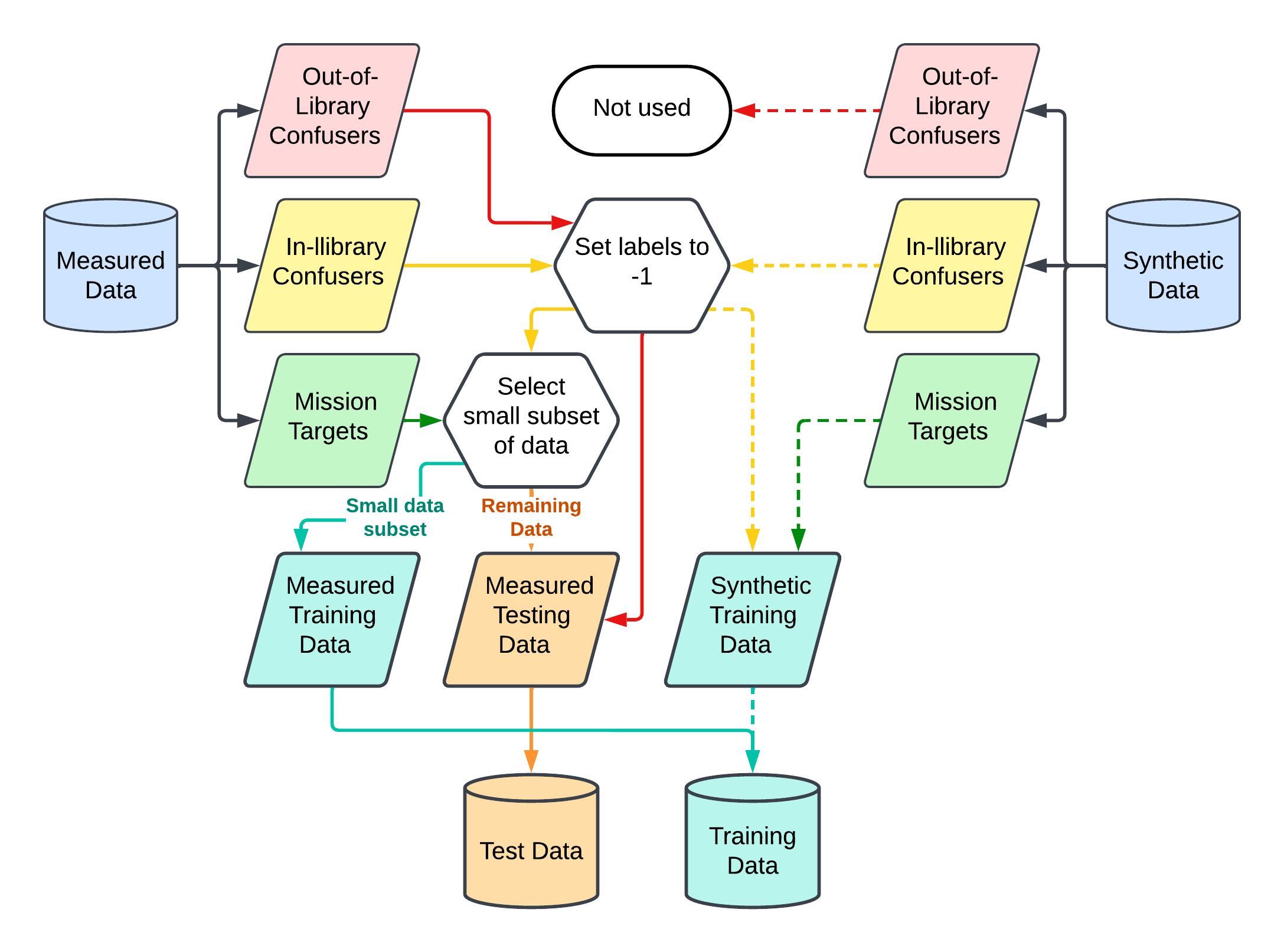}
    \caption{Dataset Preprocessing Flow}
    \label{fig:dataflow}
\end{figure}

From these different class sets, we now divide them into training, validation, and testing datasets. Note for simplicity in Figure~\ref{fig:dataflow} and in future sections we use ``training data'' to represent both training and validation data, as we use a small subset of our training data for validation purposes. For measured data, we only want to train our models on a small portion of the data and use the remainder for testing. Therefore, we select a small sample of $I$ images from our measured data with labels in $\mathcal{C}_{mt}\cup \mathcal{C}_{ic}$ and use it for training, using the remaining images for testing. This subset of measured training images can be formally defined as:
\begin{align}
    \mathcal{D}_{m,\text{train}}=&\left\lbrace \vx_i, y_i\right\rbrace _{i=1}^I\\
    \mathcal{D}_{m, \text{train}}\subseteq&\mathcal{D}_m\\
    y_i\in & \mathcal{C}_{mt}\cup \mathcal{C}_{ic};\;\forall i\in I
\end{align}
The implications of the selection of $I$ will be discussed in Section~\ref{sec:results}. Finally we use all measured data not in the training set as our test set, and can define it formally as such:
\begin{align}
    \mathcal{D}_{m, \text{test}}=\mathcal{D}_m\setminus\mathcal{D}_{m, \text{train}}
\end{align}

For synthetic data, we isolate the out-of-library confusers and then use the remaining data as training data. We do not test on synthetic data as it is in-distribution data and not the purpose of these experiments. This is formally defined as follows:
\begin{align}
    \mathcal{D}_{s, \text{train}}=&\left\lbrace \vx_k, y_k\right\rbrace_{k=1}^K\\
    \mathcal{D}_{s, \text{train}}\subseteq&\mathcal{D}_s\\
    y_k\in& C_{mt}\cup C_{ic};\;\forall k\in K\\
    \mathcal{D}_{s, \text{test}}=&\varnothing
\end{align}
where $K$ is the number of synthetic images we train on.

\textbf{We also note that each model is trained with the same mission targets, confusers, and training measured data set of $I$ images to prevent the inflation of results.} We can observe this dataset preprocessing flow as shown in Figure~\ref{fig:dataflow}.

\subsection{Model Training}

First, we need to configure the training environment that we will be ensembling in. For each model, we use ResNet-18 which has shown promise in previous SAR image classification experiments \cite{he2016deep, soldin2018sar, huang2020classification}. Each network in our ensemble is trained on the same set of data, as depicted in Figure~\ref{fig:dataflow} by ``Training Data", as well as in the same environment (same learning rate, optimizer, loss function, and epochs). The only difference between the model training environments is that all model weights are initialized randomly based on a normal Gaussian distribution.

In our experiment, we train 25 models for each value of $I\in\{0, 50, 100, 150, 200, 250\}$ measured images selected using stratified sampling to determine the number of images sampled from each class.

\subsection{Metrics}

Now we review the metrics that define the success of our experiments.

\subsubsection{AUROC}
\label{subsubsec:auroc}
The first metric is the Receiver-Operating Characteristic Curve (ROC) \cite{bradley1997use}, which plots the true positive rate (tpr), the probability of correctly flagging a confuser, against the false positive rate (fpr), the probability of mistakenly flagging a mission target as a confuser, as the feature magnitude threshold $\delta$ is varied. The Area Under the ROC (AUROC) is a value between 0 and 1 that provides a threshold-independent way of evaluating the separation between the mission targets and confusers. The closer our AUROC value is to $1$, the better our model is at differentiating between mission targets and confusers. We can also use the ROC to select a reasonable threshold value. In our experiments, we seek to maximize the ratio of true positives to false positives to calculate our threshold, $\delta$, as the cutoff between classifying a target as a mission target or a confuser. We calculate this value in our experiments as shown below:
\begin{align}
    idx=&\argmax_k (\text{tpr}[k]-\text{fpr}[k])\\
    \delta=&\text{thresholds}[idx]
\end{align}
where $idx$ represents the optimal index between the true and false positive rates and ``thresholds'' refers to the threshold values at which these true and false positive rates were observed.

\subsubsection{Accuracy}
Our second metric represents the classification success of our models through classification accuracy and confusion matrices. We consider two different methods of computing accuracy. First, we define the overall accuracy, conditioned on the feature magnitude threshold $\delta$ defined in Section~\ref{subsubsec:auroc}, as the accuracy computed across all classes, where confusers are assigned to an extra class. However, to evaluate mission target feature separation in a threshold-independent manner, we also consider the classification accuracy of mission targets only. Taken together with AUROC, this provides threshold-independent measures of feature separation. By observing the confusion matrices and mission target accuracy, we are able to see if there is an improvement in confuser rejection while still maintaining (or improving upon) the correct classification of mission targets.

\subsection{Results}
\label{sec:results}

In this section, we review the performance of our methods through the our conducted experiments.

\subsubsection{Baseline Performance}

\begin{figure}[h]
    \centering
    \begin{subfigure}{0.4\textwidth}
        \includegraphics[width=1\textwidth]{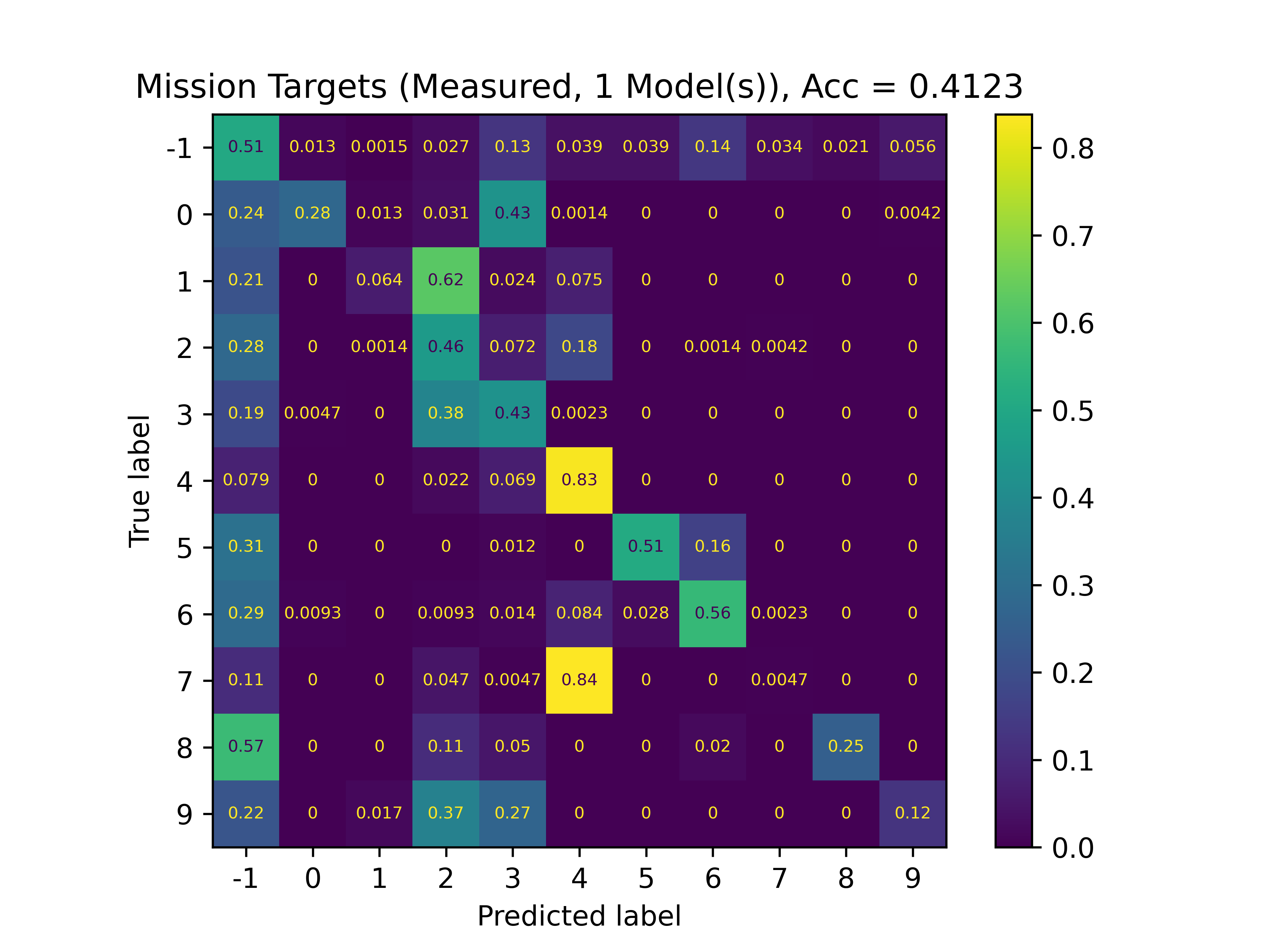}
        \caption{$I=0$}
    \end{subfigure}
    \begin{subfigure}{0.4\textwidth}
        \includegraphics[width=\textwidth]{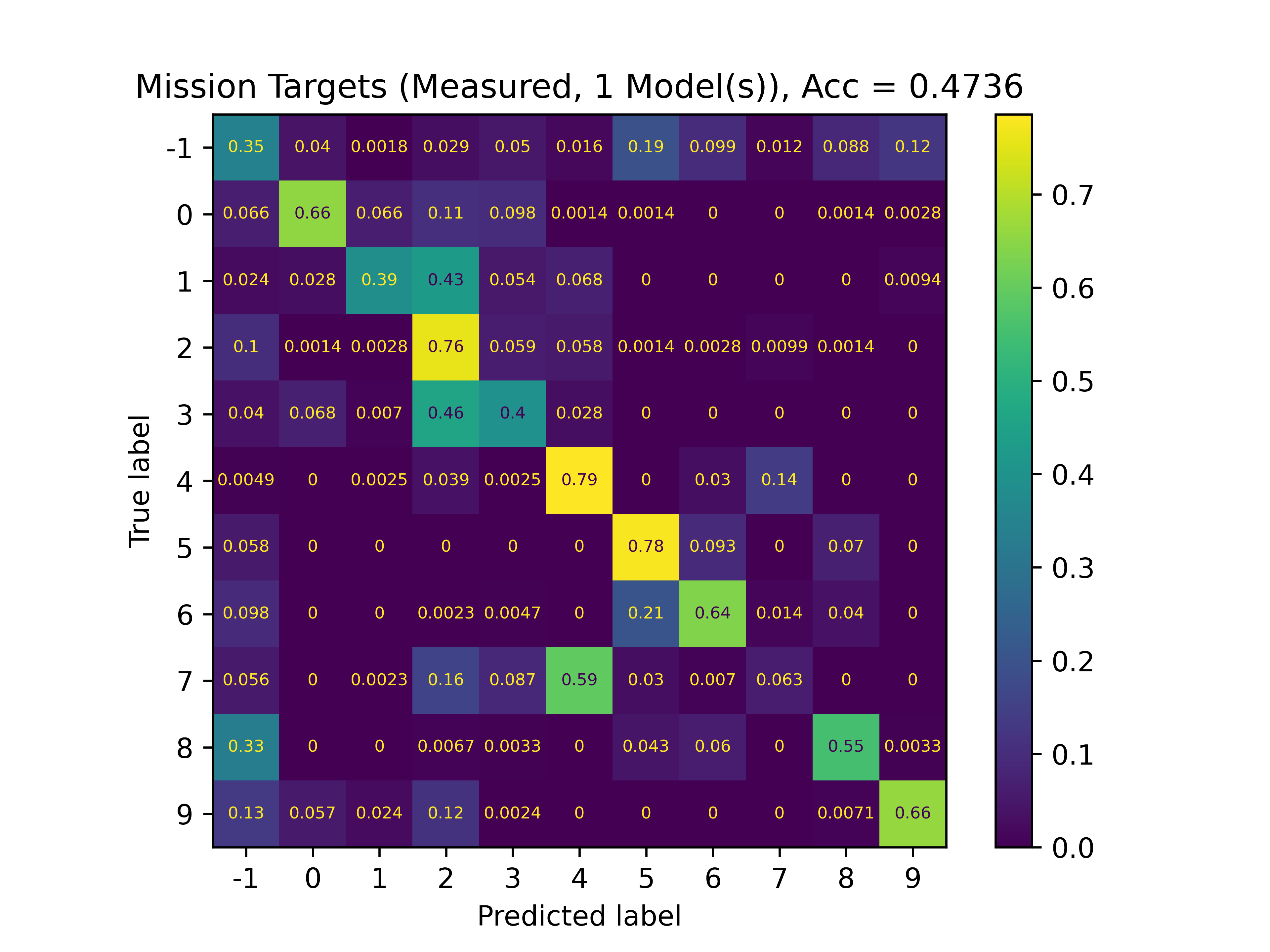}
        \caption{$I=50$}
    \end{subfigure}
    \begin{subfigure}{0.4\textwidth}
        \includegraphics[width=1\textwidth]{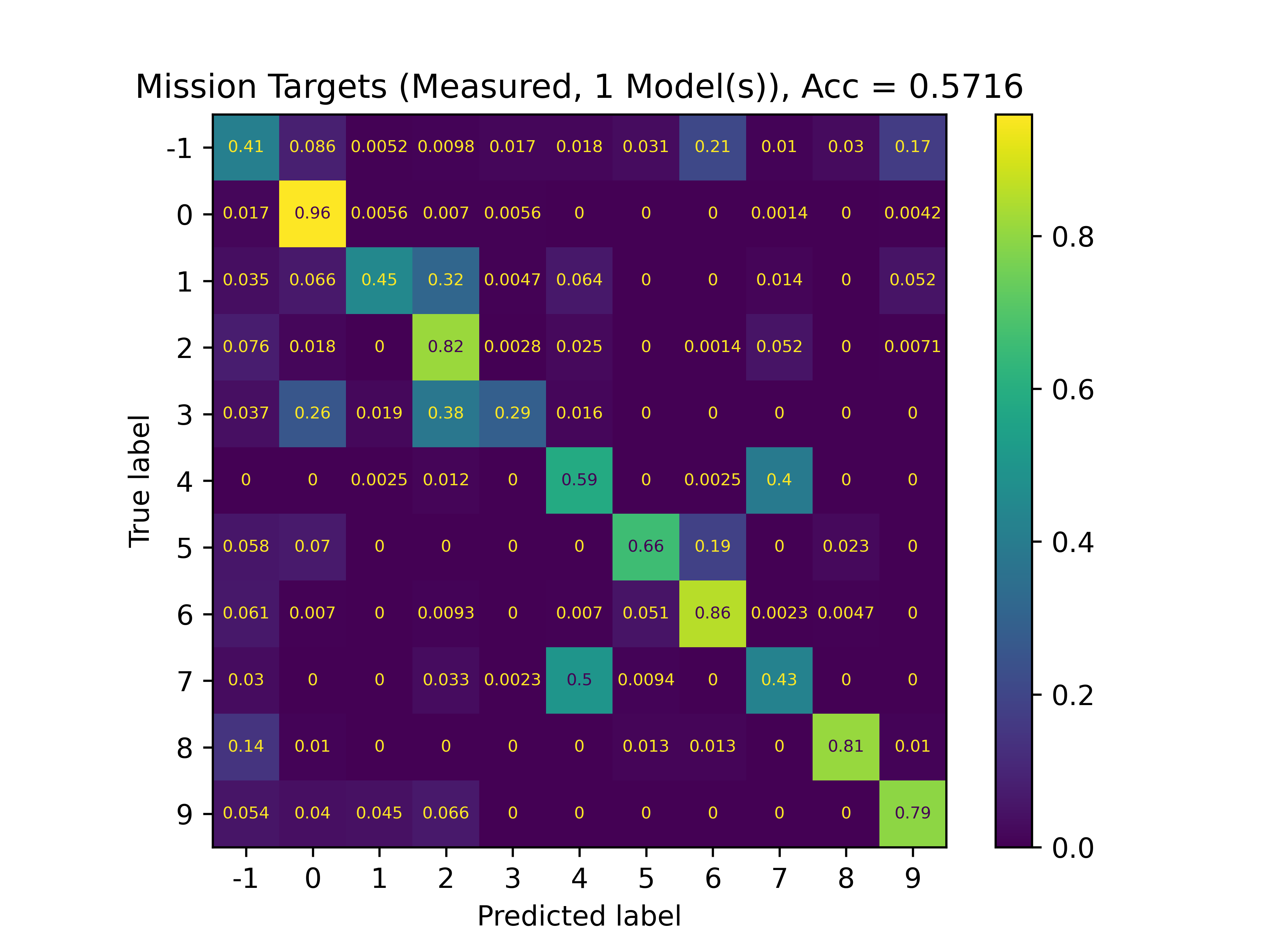}
        \caption{$I=100$}
    \end{subfigure}
    \begin{subfigure}{0.4\textwidth}
        \includegraphics[width=\textwidth]{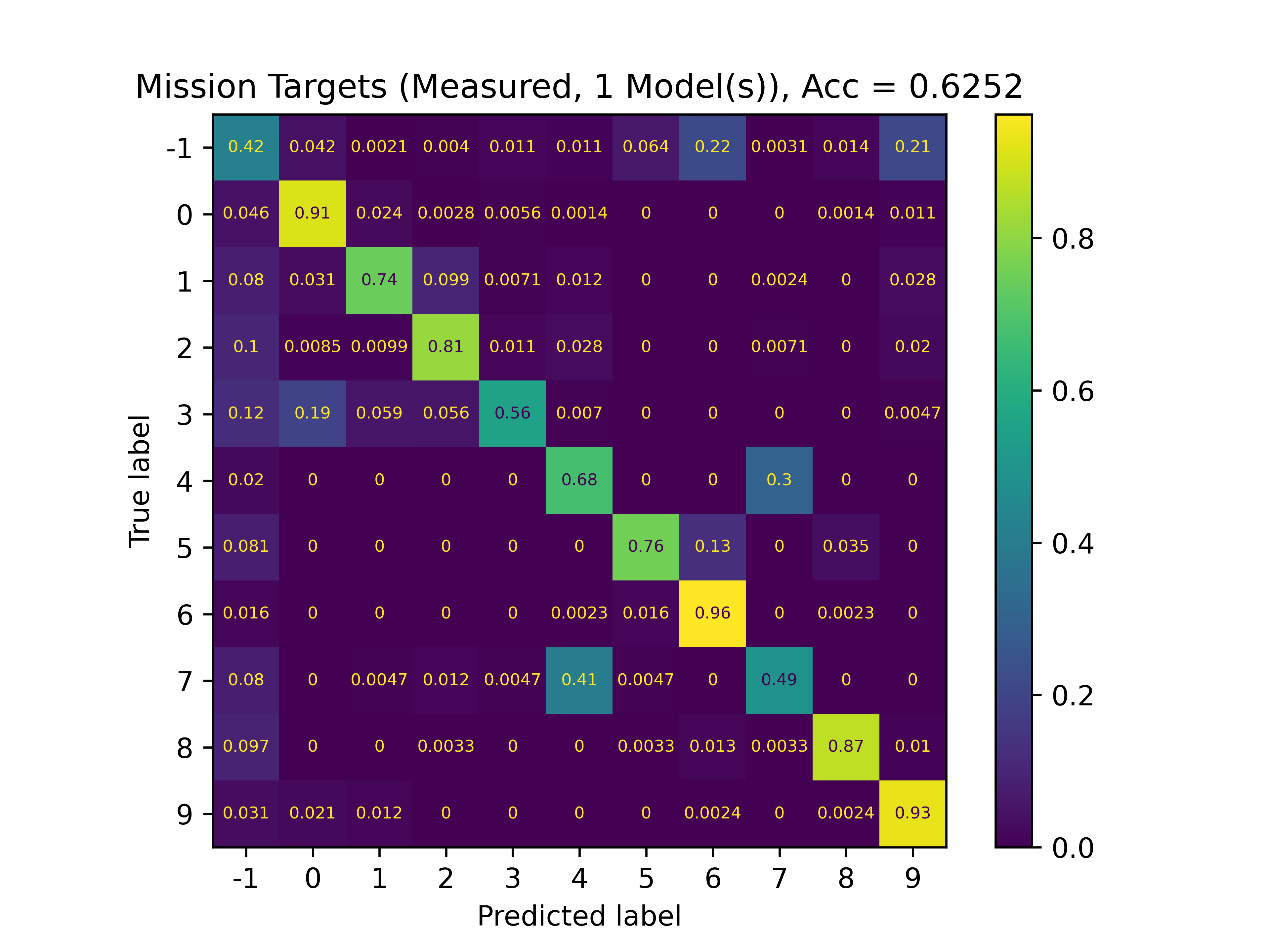}
        \caption{$I=150$}
    \end{subfigure}
    \begin{subfigure}{0.4\textwidth}
        \includegraphics[width=\textwidth]{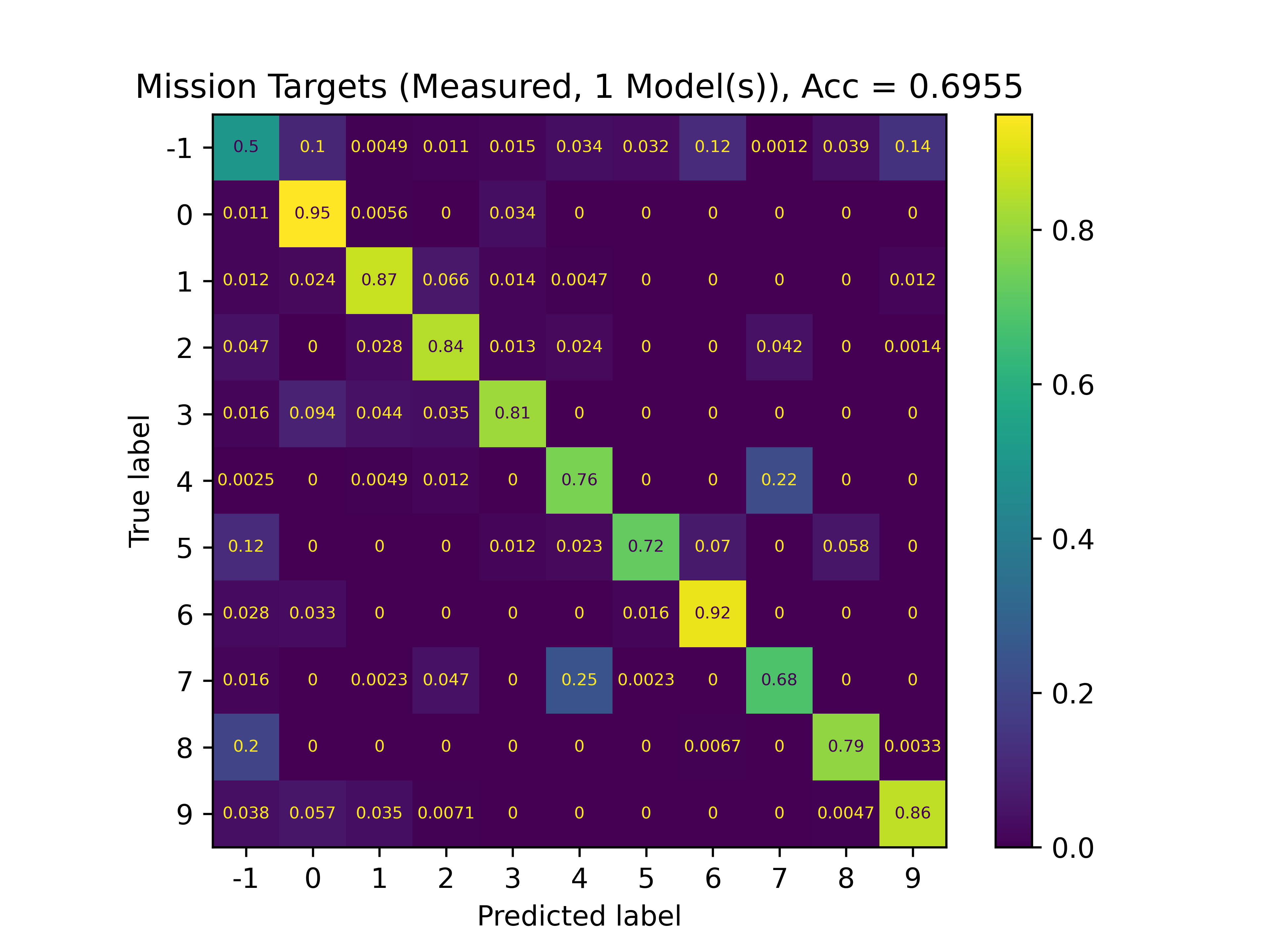}
        \caption{$I=200$}
    \end{subfigure}
    \begin{subfigure}{0.4\textwidth}
        \includegraphics[width=\textwidth]{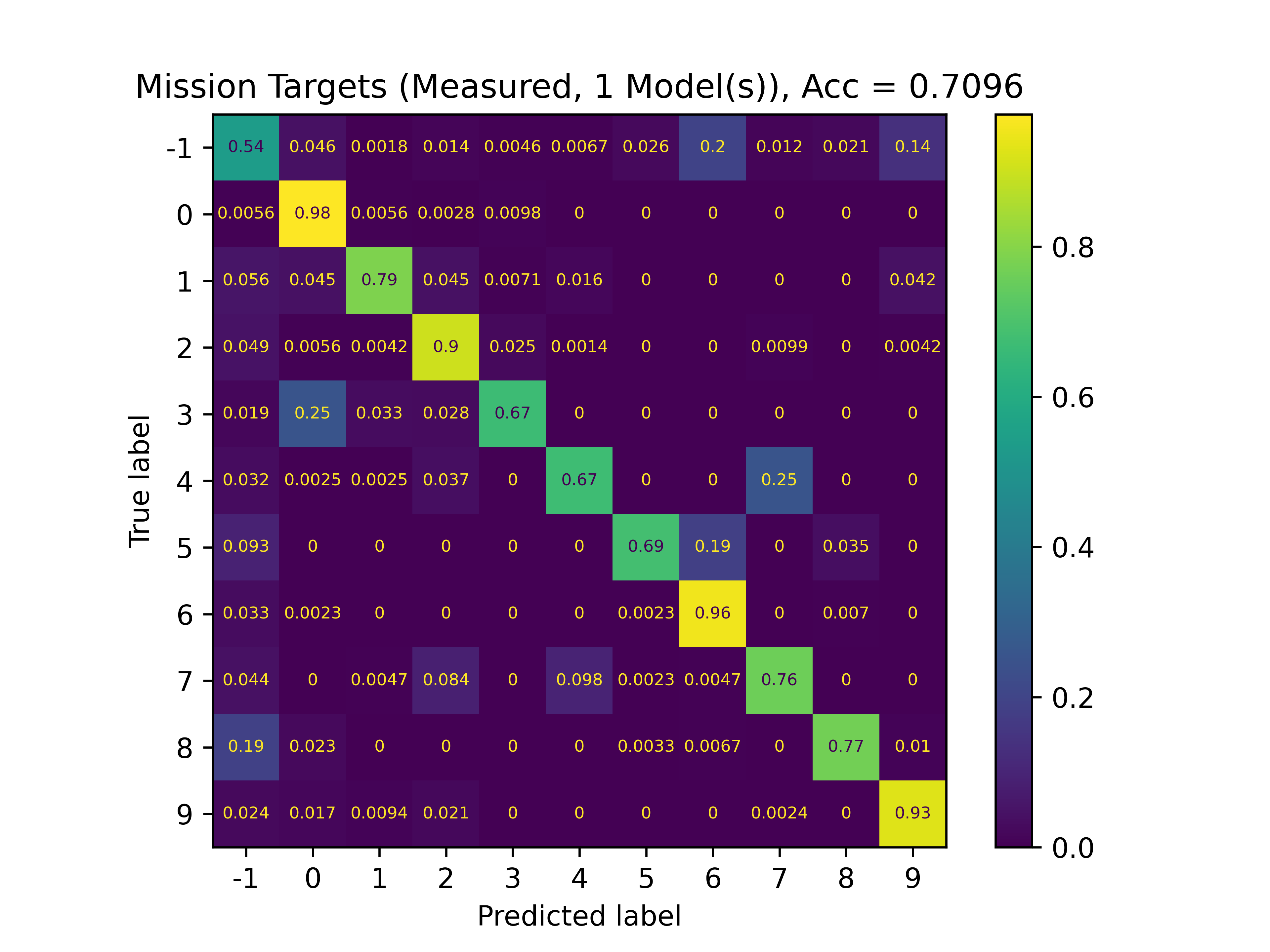}
        \caption{$I=250$}
    \end{subfigure}
    \caption{Baseline: Singular model confusion matrices trained on different amounts of measured data where confusers are detected with threshold $\delta$, as discussed in Section~\ref{subsubsec:ensm_overview} and given the label $-1$.}
    \label{fig:baseline_conf_mat}
\end{figure}

To show the efficacy of our various techniques, we first need to establish a baseline performance. This baseline consists of the results from a singular trained model from each value of $I\in\{0, 50, 100, 150, 200, 250\}$ measured training samples, as shown in Figure~\ref{fig:baseline_conf_mat}.

\begin{figure}
    \centering
    \begin{subfigure}{0.4\textwidth}
        \includegraphics[width=1\textwidth]{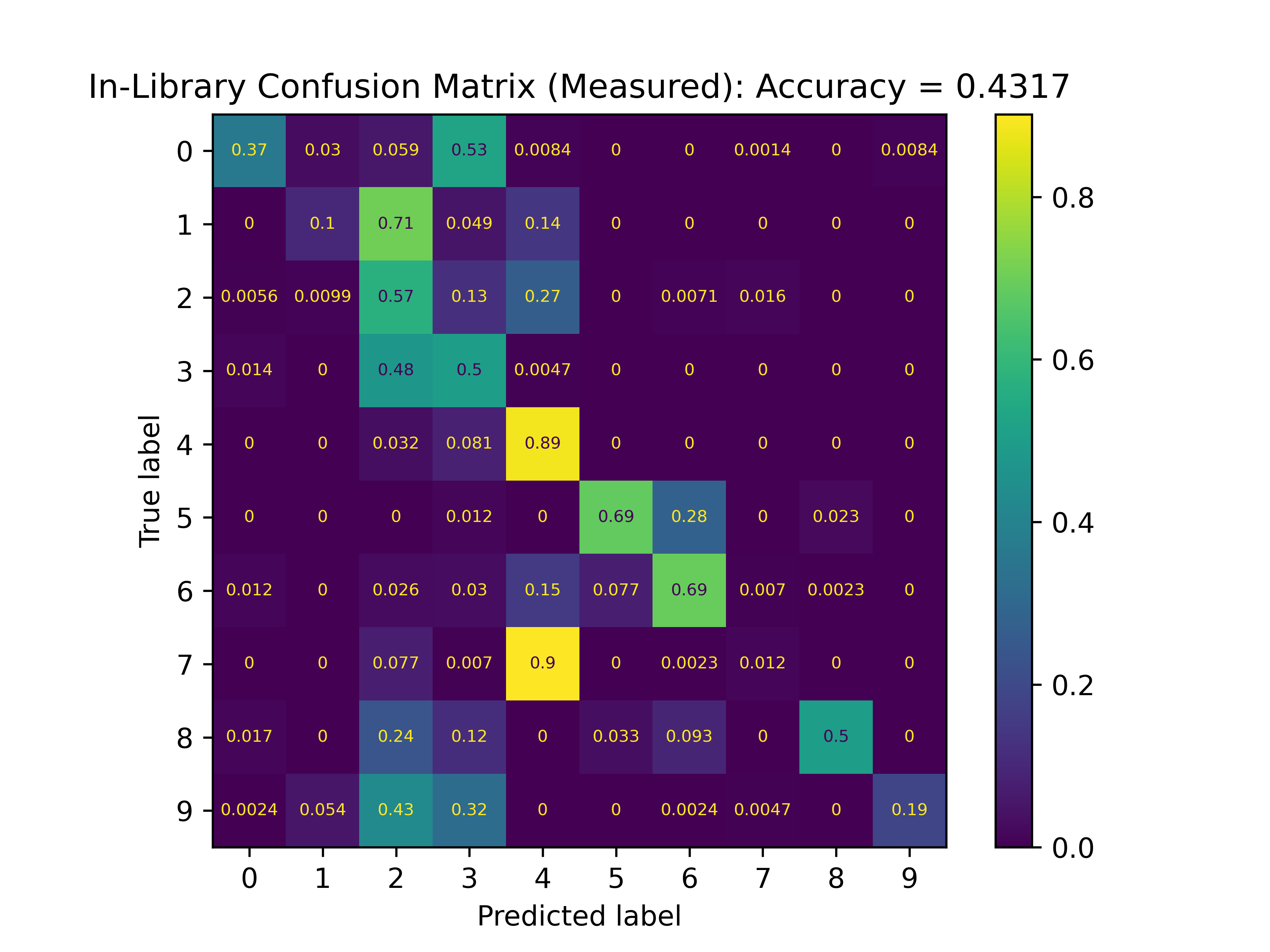}
        \caption{$I=0$}
    \end{subfigure}
    \begin{subfigure}{0.4\textwidth}
        \includegraphics[width=\textwidth]{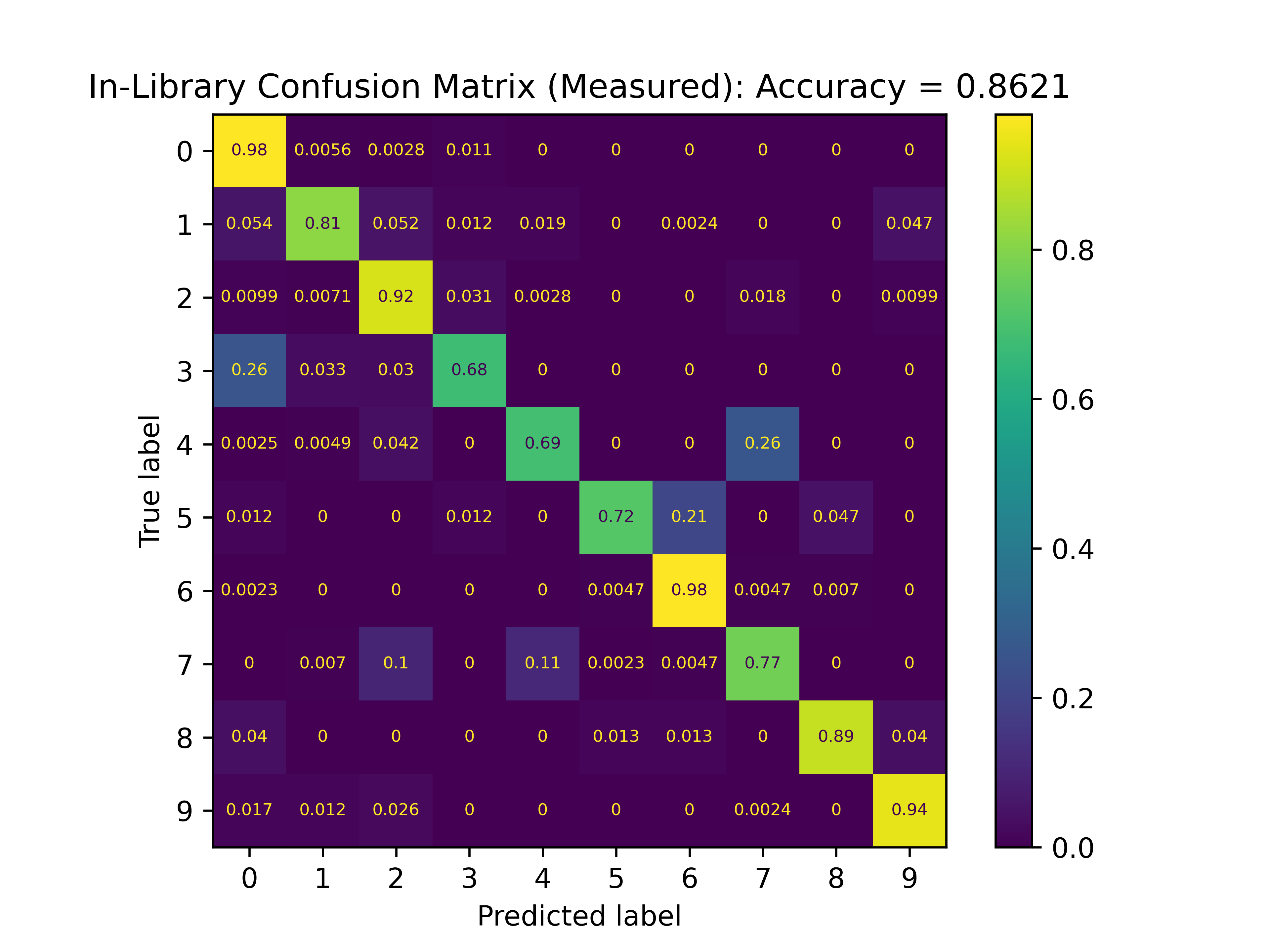}
        \caption{$I=250$}
    \end{subfigure}
    \caption{Baseline: Singular model confusion matrices trained on different amounts of measured data where confusers are withheld and the threshold-independent accuracy is computed over mission targets only.}
    \label{fig:baseline_conf_mat_noconf}
\end{figure}

From this baseline measurement alone, we can see that the use of minimal measured data as a supplement to the synthetic data has a positive correlation on overall classification accuracy, as the accuracy goes from $0.4123\rightarrow 0.7096$ by only using a small subset of the measured dataset. This improvement is also reflected in the threshold-independent mission-target accuracy, as shown in Figure~\ref{fig:baseline_conf_mat_noconf} ($0.4317\rightarrow 0.8621$). However, this is expected, as we are supplementing our training set with data from the original distribution and not the synthetically generated one. Furthermore, we can see that this technique has diminishing effects the more measured data we introduce into the training set. This is beneficial, however, as that means we only need minimal amounts of measured data to train a strong classification model.

Finally, we can observe the behavior of the model when trained on just synthetically generated data. It appears to have strong bias towards several classes when training, which can be subject to many things including a lack of accurately generated data for certain targets or similarity in target features causing heightened misclassification.

\subsubsection{Unweighted Model Averaging}

\begin{figure}
    \centering
    \begin{subfigure}{0.45\textwidth}
        \includegraphics[width=\textwidth]{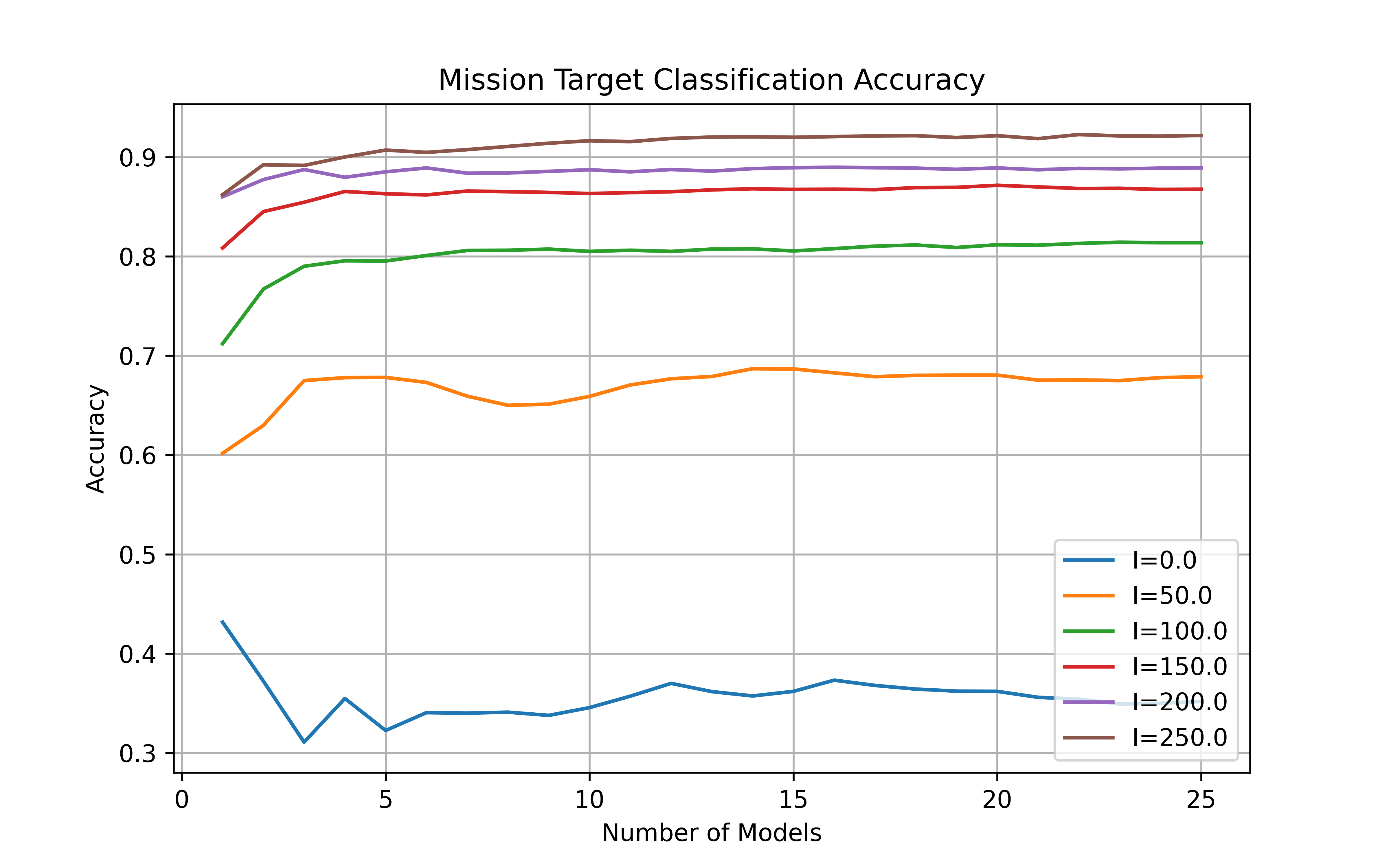}
        \caption{Mission Target Classification Accuracy}
        \label{fig:uma_mt}
    \end{subfigure}
    \begin{subfigure}{0.45\textwidth}
        \includegraphics[width=\textwidth]{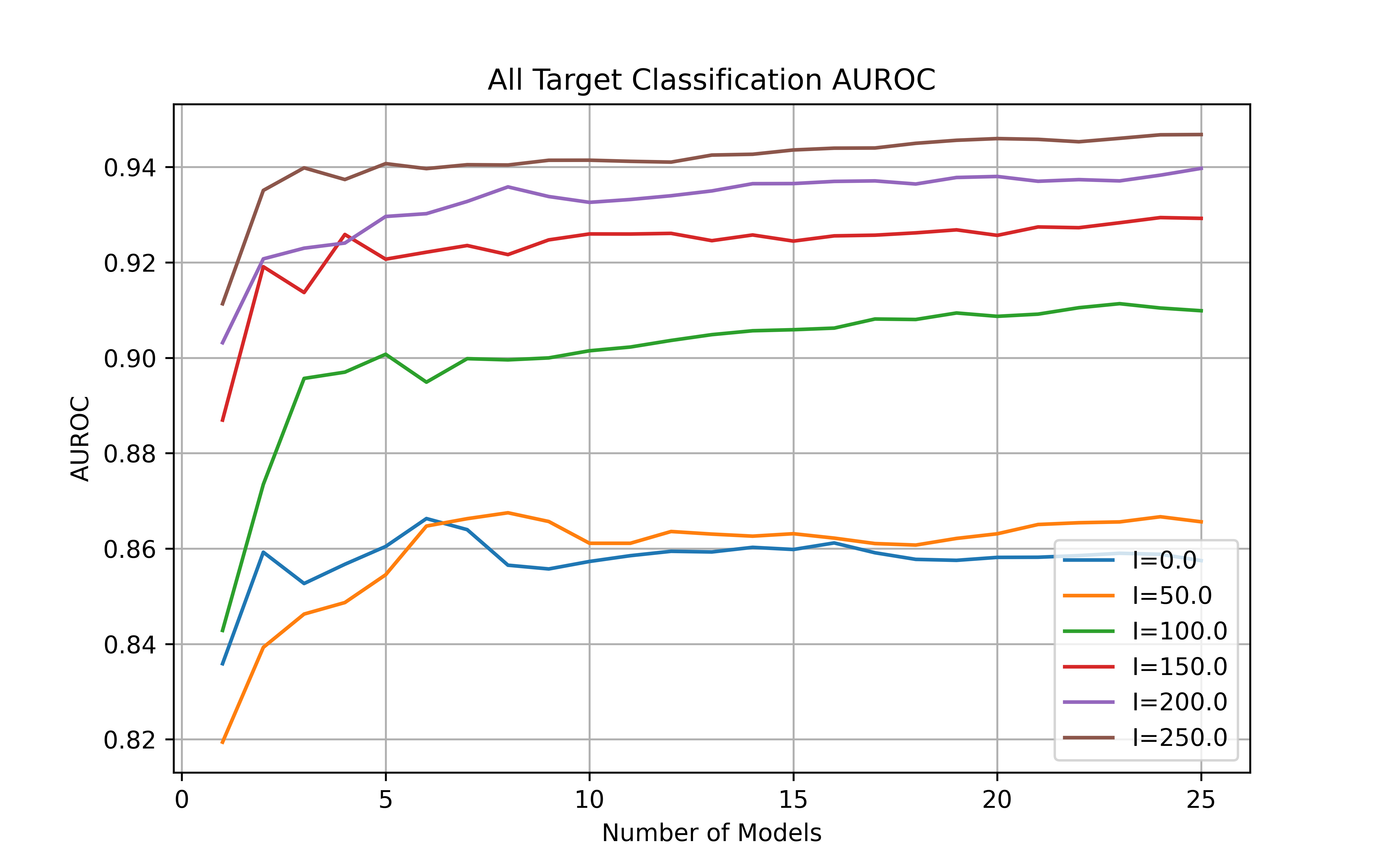}
        \caption{Known/Unknown AUROC Values}
        \label{fig:uma_c}
    \end{subfigure}
    \caption{Classification metrics using Unweighted Model Averaging on different amounts of measured data.}
    \label{fig:class_comb}    
\end{figure}

Now that we have established a baseline performance for classifications using different amounts of measured data and a singular model, we can test the effectiveness of introducing ensembling. First, we start with Unweighted Model Averaging as depicted in Section~\ref{subsec:uma}. We consider improvement as a function of the number of models in our ensemble using the threshold-independent measures of mission target accuracy and mission target/confuser AUROC, shown in Figure~\ref{fig:class_comb}.

We can make several observations from the performance of mission target classification through ensembling. First is in regards to accuracy, with there being a sharp jump in performance after ensembling just a couple models trained on measured data. We note that the performance drops in ensembling models trained with no measured data, signifying that uninformed classification only worsens with more models due to further randomness in the outputs. We also note that there are diminishing returns on accuracy after a couple models are ensembled, with experiments using a higher $I$ experiencing this effect faster than those with a lower $I$.

As for the AUROC metric, we notice that ensembling has a positive impact on all values of $I$ as shown in Figure~\ref{fig:uma_c}. While the effectiveness diminishes after a couple models, it still steadily increases with even more models added, showing that there could be even further benefit by adding in more models to reduce the classification of confusers.

Figure~\ref{fig:class_comb} gives a clear visual that ensembling has a strong positive effect on both mission target classification as well as for confuser rejection.

\begin{figure}
    \centering
    \begin{subfigure}{0.4\textwidth}
        \includegraphics[width=1\textwidth]{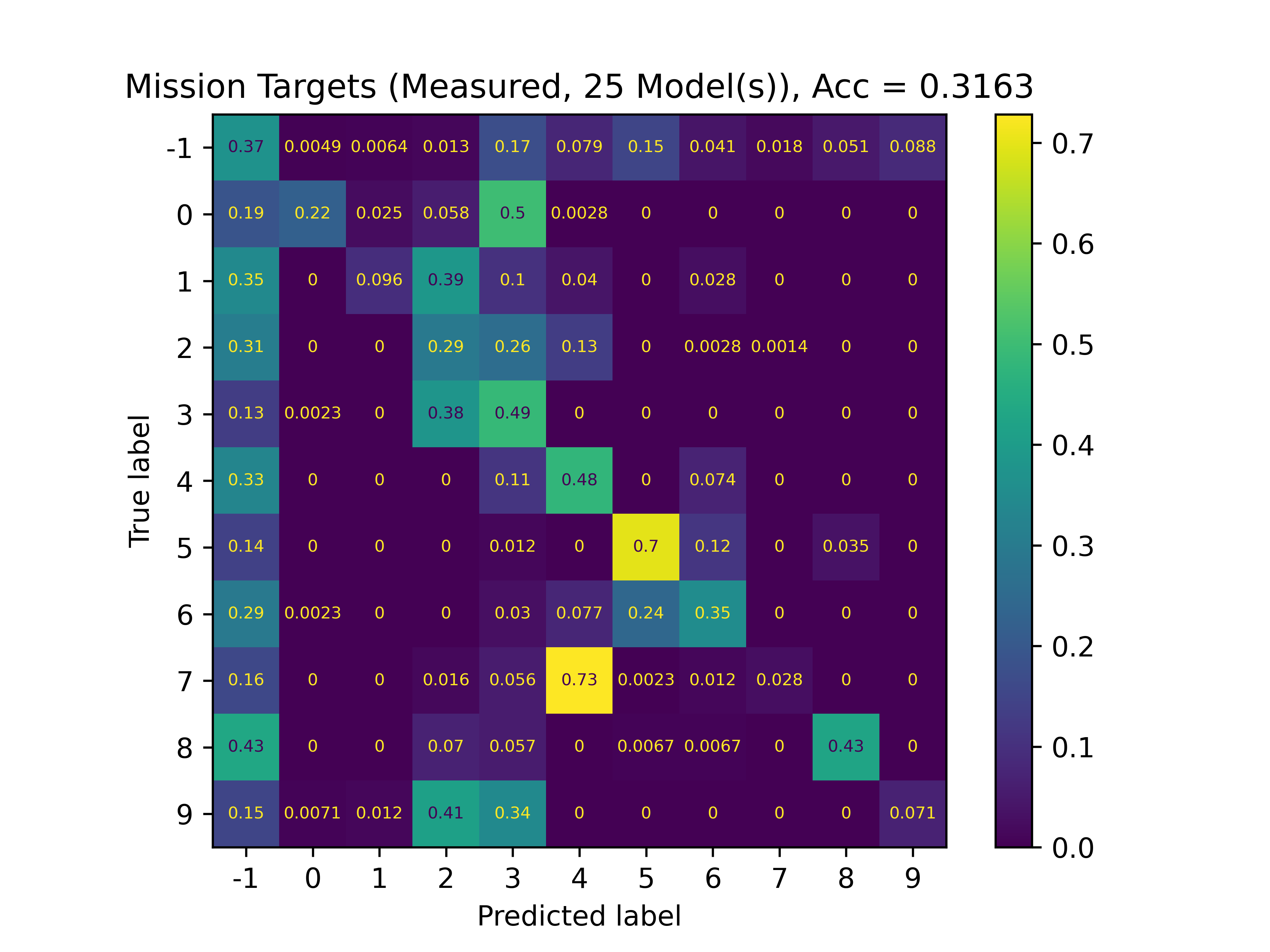}
        \caption{$I=0$}
    \end{subfigure}
    \begin{subfigure}{0.4\textwidth}
        \includegraphics[width=\textwidth]{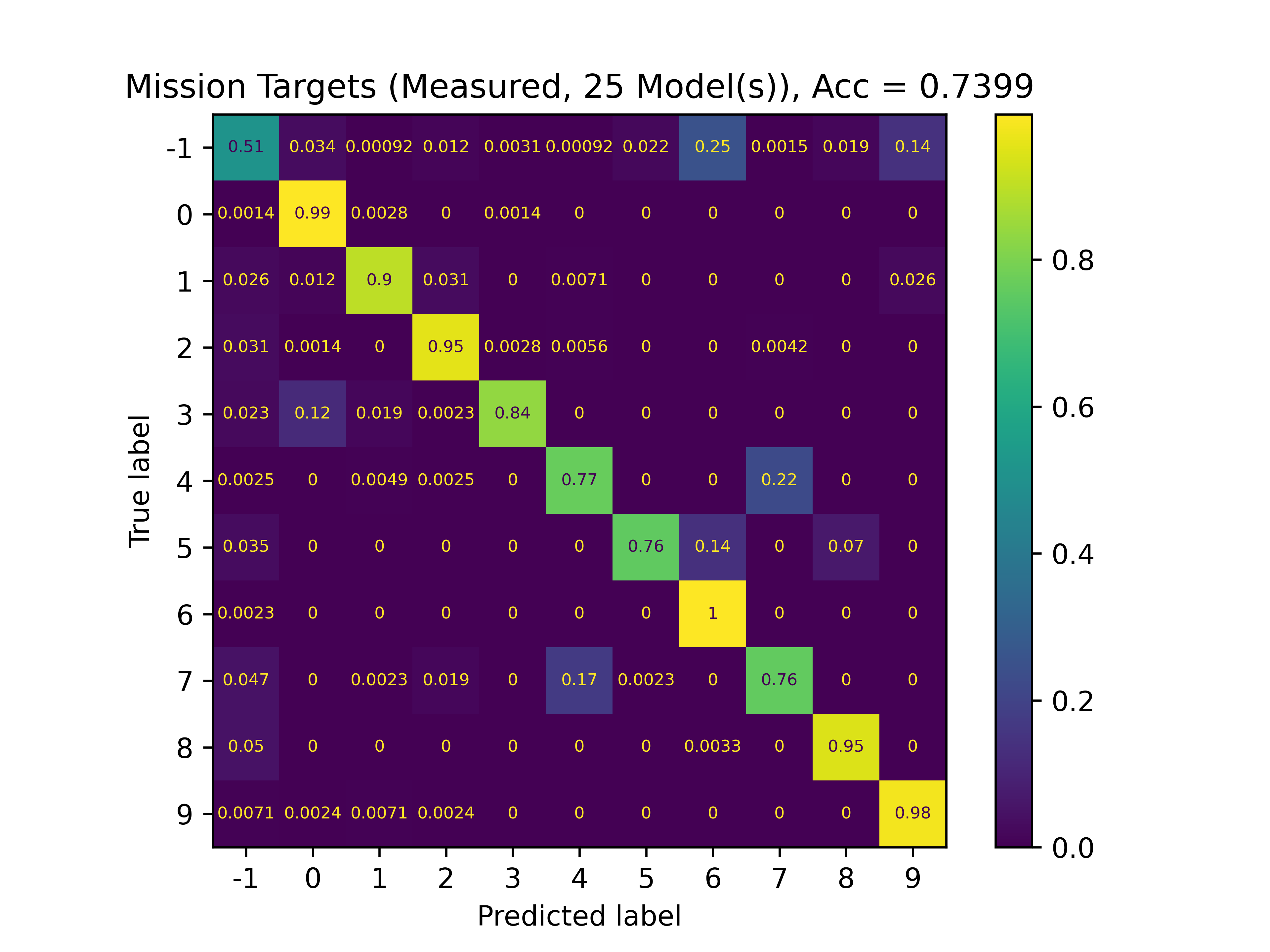}
        \caption{$I=250$}
    \end{subfigure}
    \caption{Confusion matrix performance evaluation of Unweighted Model Averaging on $25$ models trained with $I=0,250$ measured images.}
    \label{fig:uma_c_conf}
\end{figure}

Finally, we observe the overall accuracy given our chosen feature magnitude threshold as shown in Figure~\ref{fig:uma_c_conf}. The overall accuracy is lower than the mission target accuracy given mistakes made by the confuser rejection thresholding. In particular, for this choice of threshold $\delta$, the overall accuracy is reduced by a large number of false negatives (confusers are misclassified by the model as classes 6 and 9).

\subsubsection{Weighted Model Calibration}
\label{ssubsec:results_wmc}

\begin{figure}[hbt]
    \centering
    \begin{subfigure}{0.4\textwidth}
        \includegraphics[width=1\textwidth]{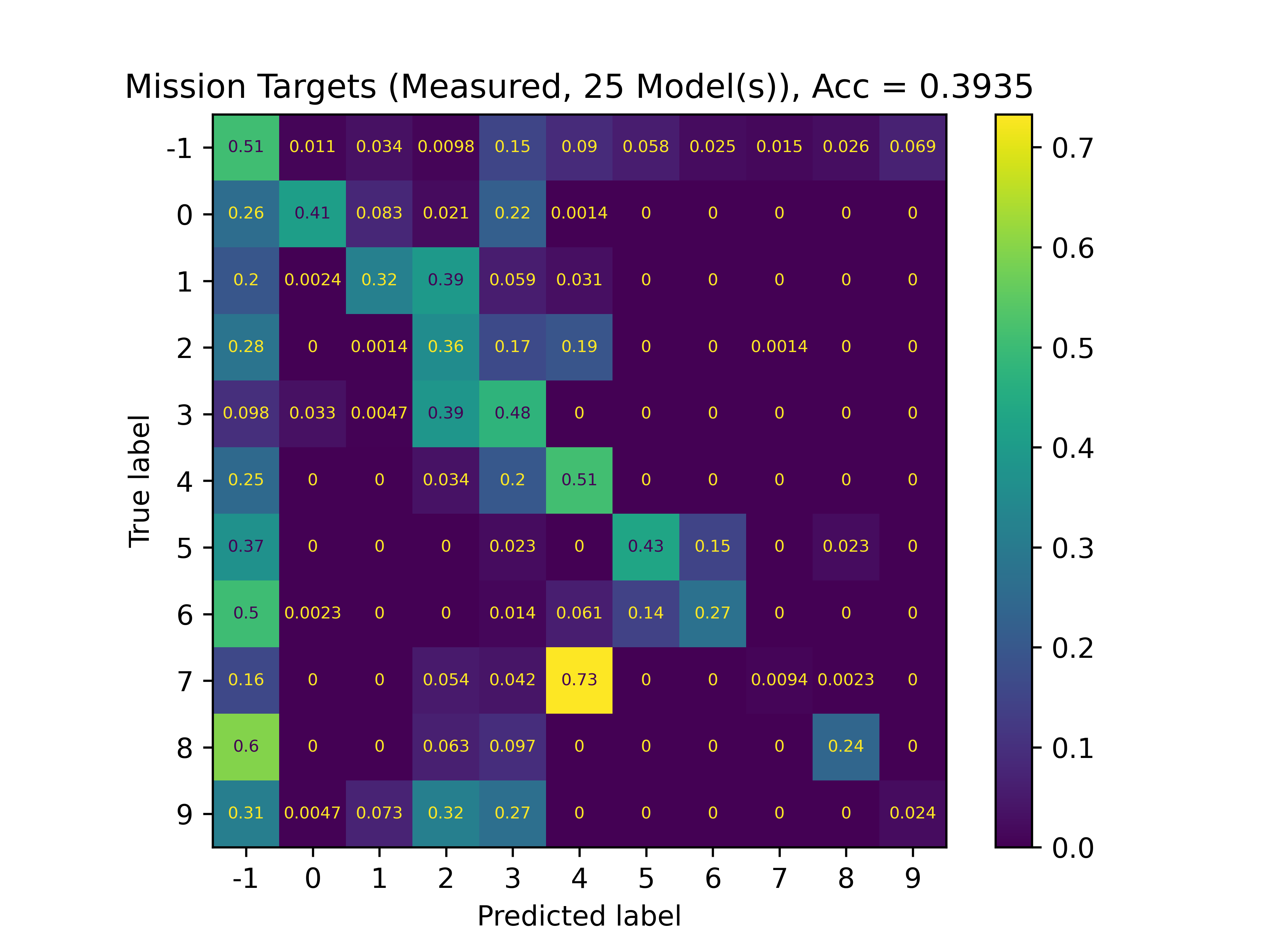}
        \caption{$I=0$}
    \end{subfigure}
    \begin{subfigure}{0.4\textwidth}
        \includegraphics[width=\textwidth]{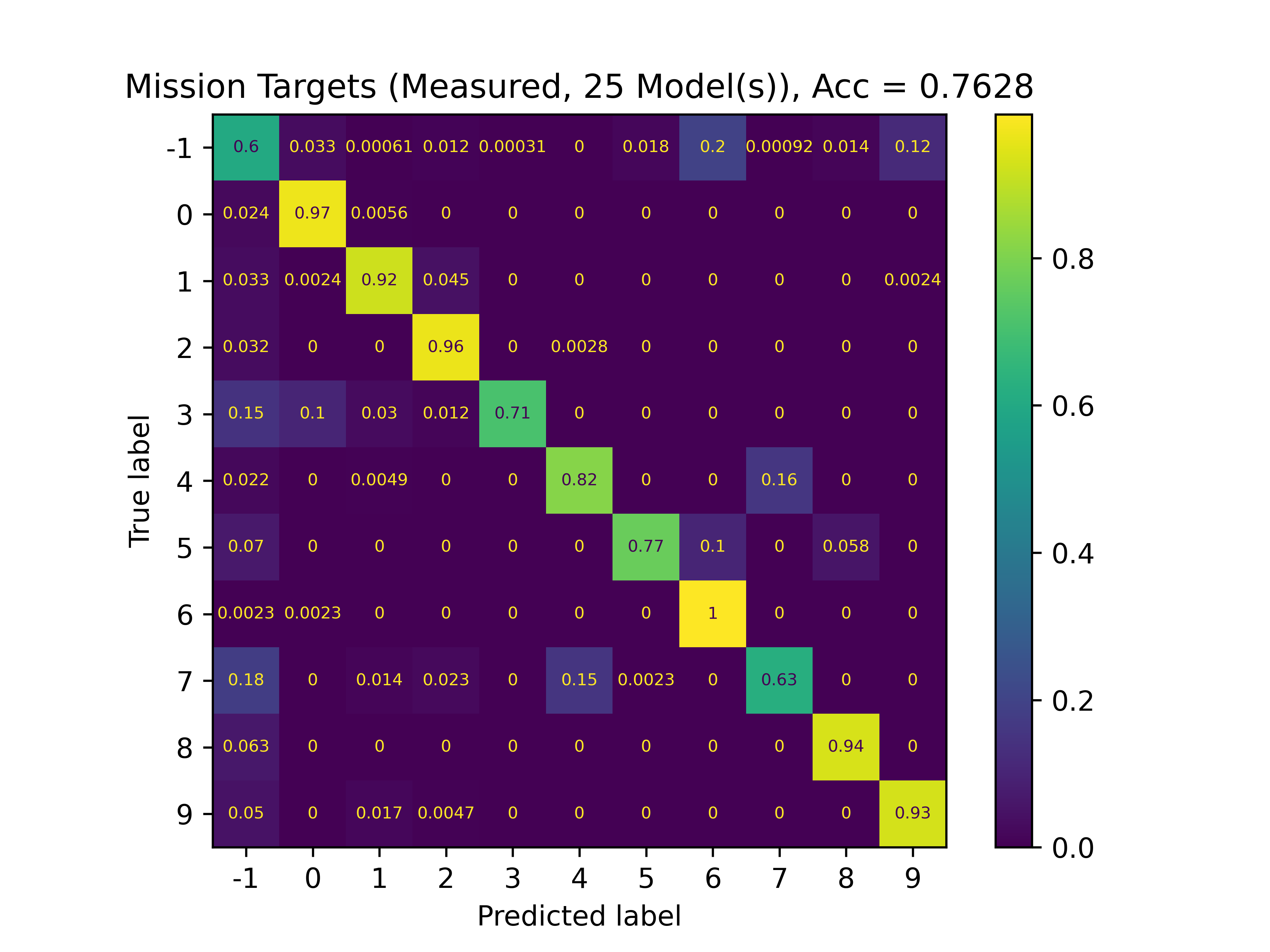}
        \caption{$I=250$}
    \end{subfigure}
    \caption{Confusion matrix performance evaluation of Weighted Model Calibration on $25$ models trained with $I=0,250$ measured images}
    \label{fig:wmc_c_conf}
\end{figure}

Next we observe the performance of Weighted Model Calibration as described in Section~\ref{subsec:wmc}. Note that we only trained the calibration layer on $M=25$ as we assume that performance only increases the more models we calibrate on based on our results for Unweighted Model Averaging. To show the efficacy of this metric, observe the following confusion matrices for $I=0,250$ in Figure~\ref{fig:wmc_c_conf}. From these plots we can notice that there is an improvement in performance when compared to Unweighted Model Averaging for both the $I=0$ and $I=250$ cases. We also note that this technique is able to classify $10\%$ of the confusers correctly over Unweighted Model Averaging, which is most likely the main contribution to its performance increase.

Since we only tested on $M=25$, we do not have a plot comparing the AUROC values over different $M$, however, we can note the AUROC values to be $0.813$ for $I=0$ and $0.951$ for $I=250$. These values are relatively similar to the Unweighted Model Averaging technique and will be further explored in Section~\ref{subsec:featmag} and can be seen in Table~\ref{tbl:overview}.

\subsubsection{Feature Magnitude}
\label{subsec:featmag}

\begin{figure}[h]
    \centering
    \begin{subfigure}{0.4\textwidth}
        \includegraphics[width=1\textwidth]{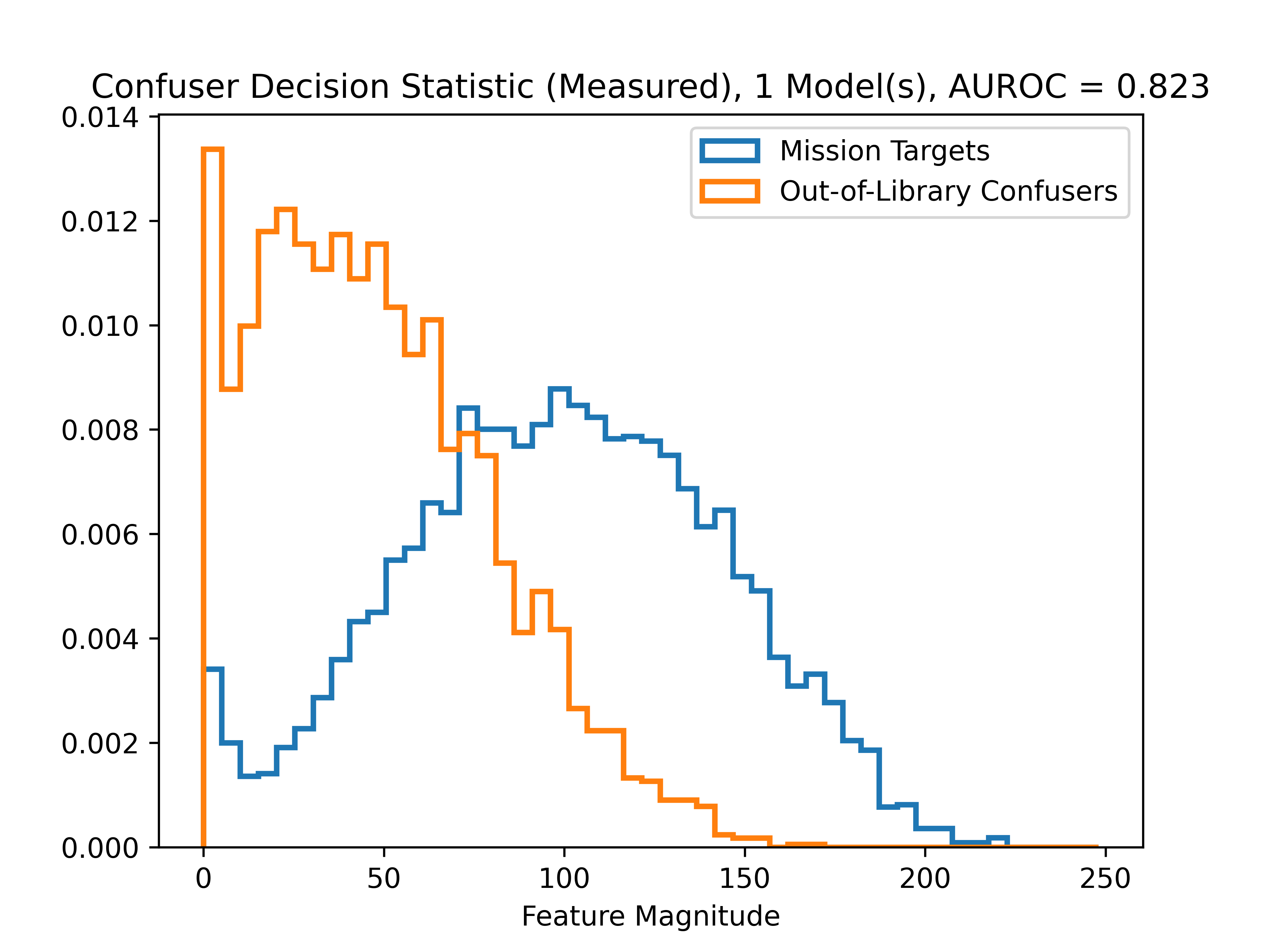}
        \caption{$I=0$}
    \end{subfigure}
    \begin{subfigure}{0.4\textwidth}
        \includegraphics[width=\textwidth]{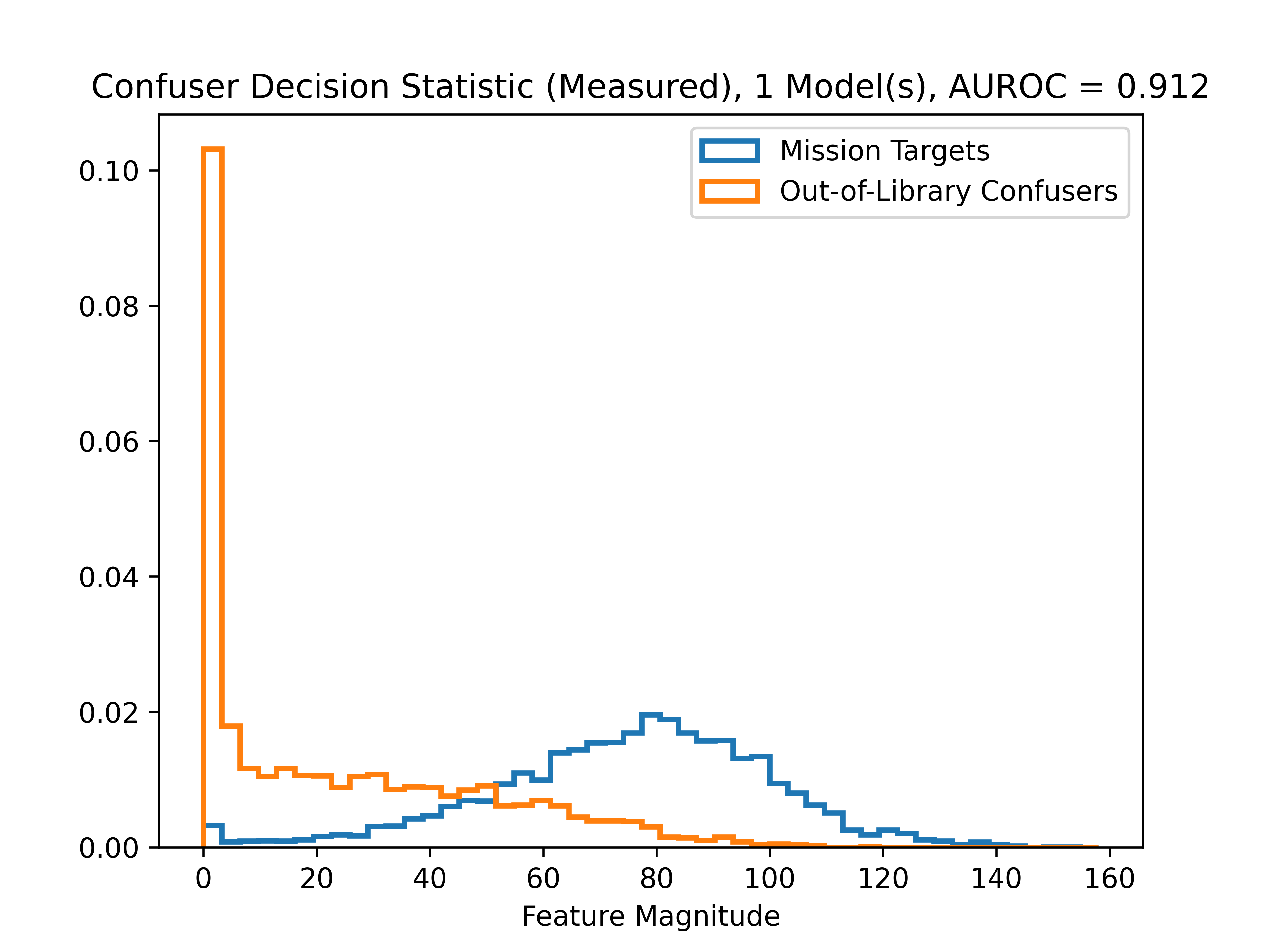}
        \caption{$I=250$}
    \end{subfigure}
    \caption{Histograms comparing the feature magnitudes of mission targets to the magnitudes of out-of-library confusers for a single model trained using Objectosphere, and $I=0$ or $I=250$ measured training samples.}
    \label{fig:fm_base}
\end{figure}

Finally, we visualize the impact of ensembling models trained with the Objectosphere loss on the feature magnitude decision statistic described in Section~\ref{ssubsec:fm} and further influenced by the Objectosphere Loss as depicted in Section~\ref{subsec:lf} \cite{dhamija2018reducing}. To see the improvement, we first observe the baseline performance using only a singular model ($M=1$) on $I=0,250$ as shown in Figure~\ref{fig:fm_base}.

We can see a significant overlap when $I=0$, meaning that introduction of a threshold would yield a significant amount of false confuser classifications. Furthermore, for $I=250$ we see that the curves are slightly more separated and flattened, however, there still is a prominent overlap yielding similar results as $I=0$.

\begin{figure}[h]
    \centering
     \begin{subfigure}{0.4\textwidth}
        \includegraphics[width=1\textwidth]{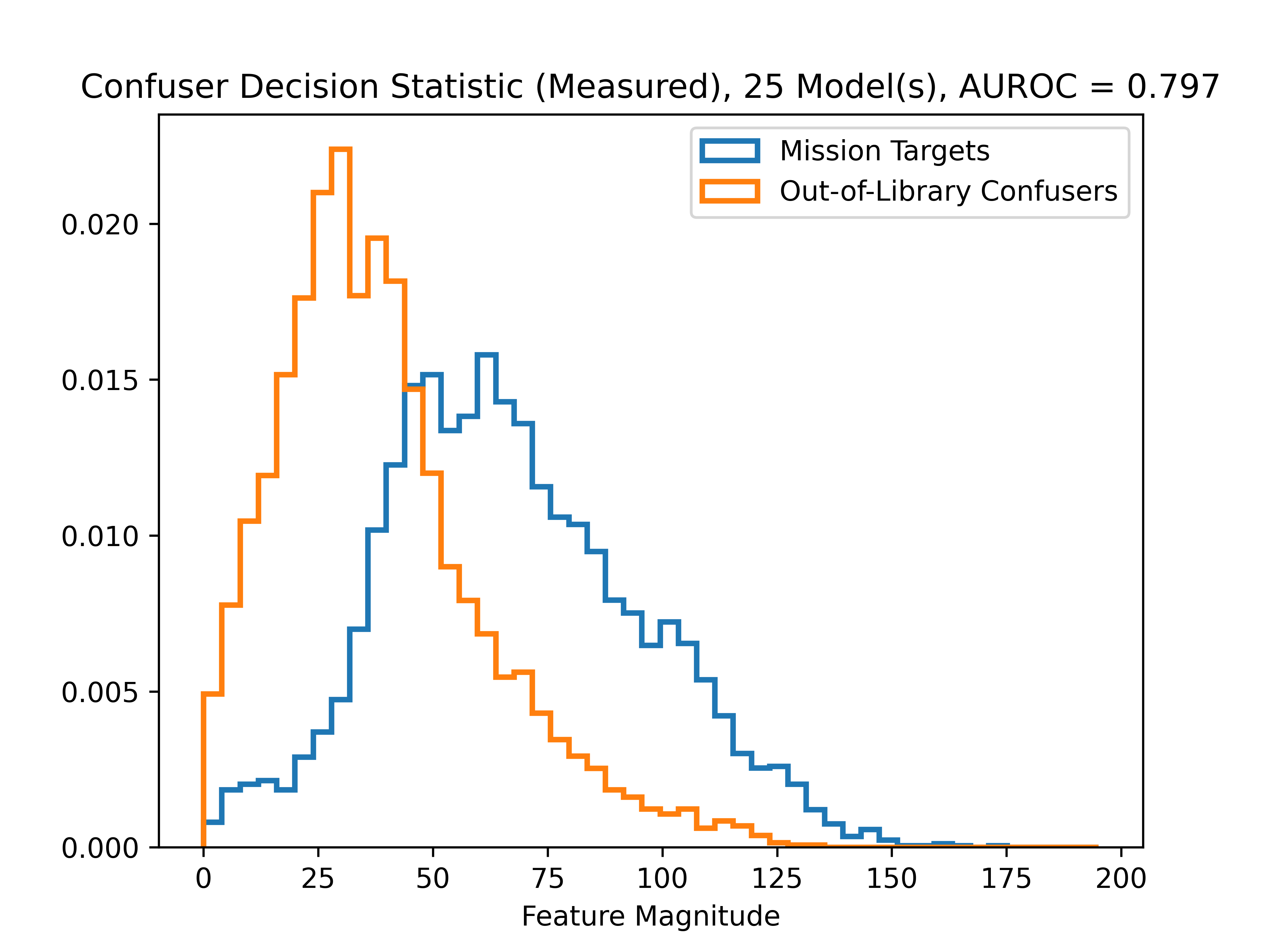}
        \caption{Unweighted Model Averaging: $I=0$}
    \end{subfigure}
    \begin{subfigure}{0.4\textwidth}
        \includegraphics[width=\textwidth]{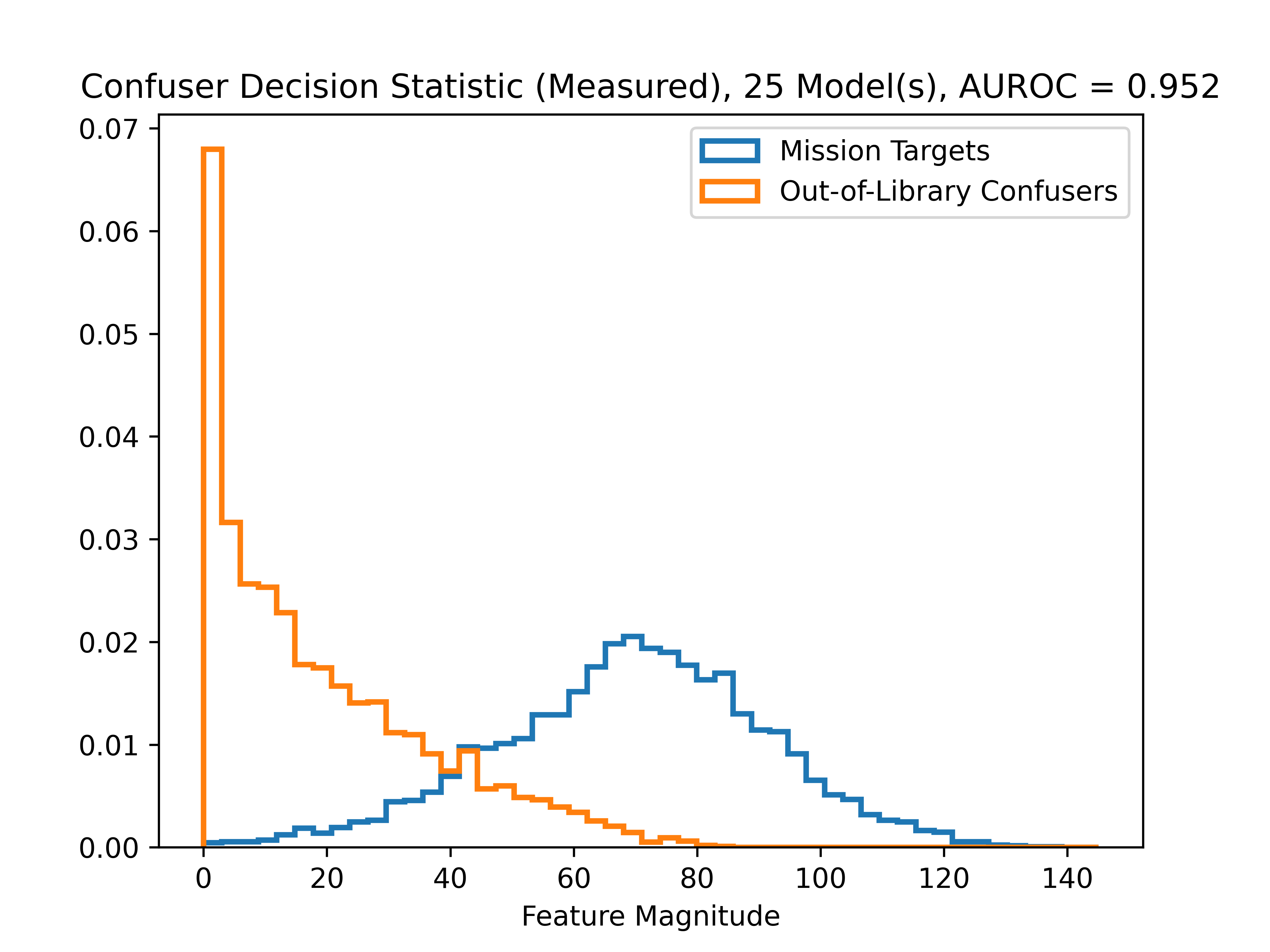}
        \caption{Unweighted Model Averaging: $I=250$}
    \end{subfigure}
    \begin{subfigure}{0.4\textwidth}
        \includegraphics[width=1\textwidth]{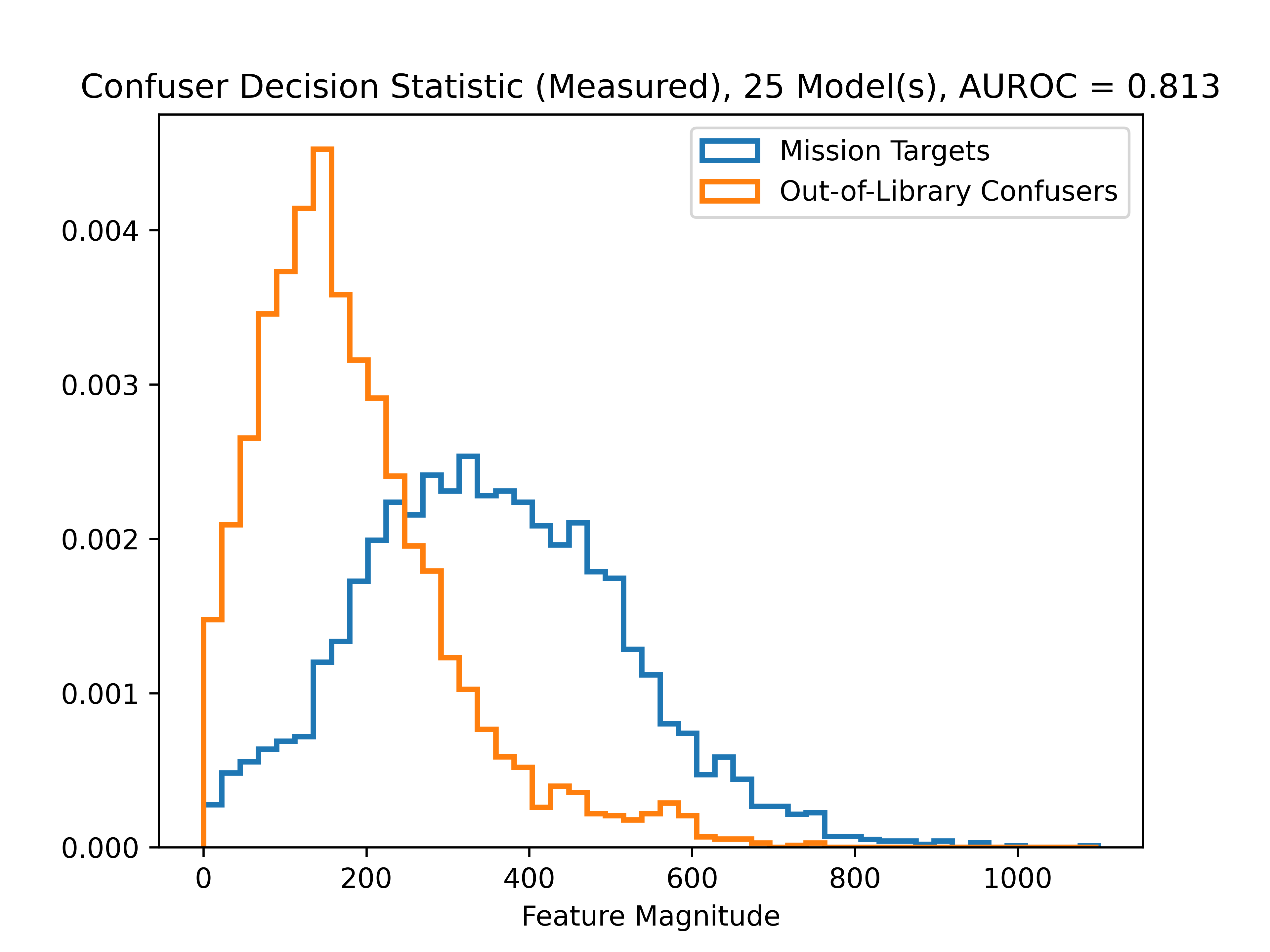}
        \caption{Weighted Model Calibration: $I=0$}
    \end{subfigure}
    \begin{subfigure}{0.4\textwidth}
        \includegraphics[width=\textwidth]{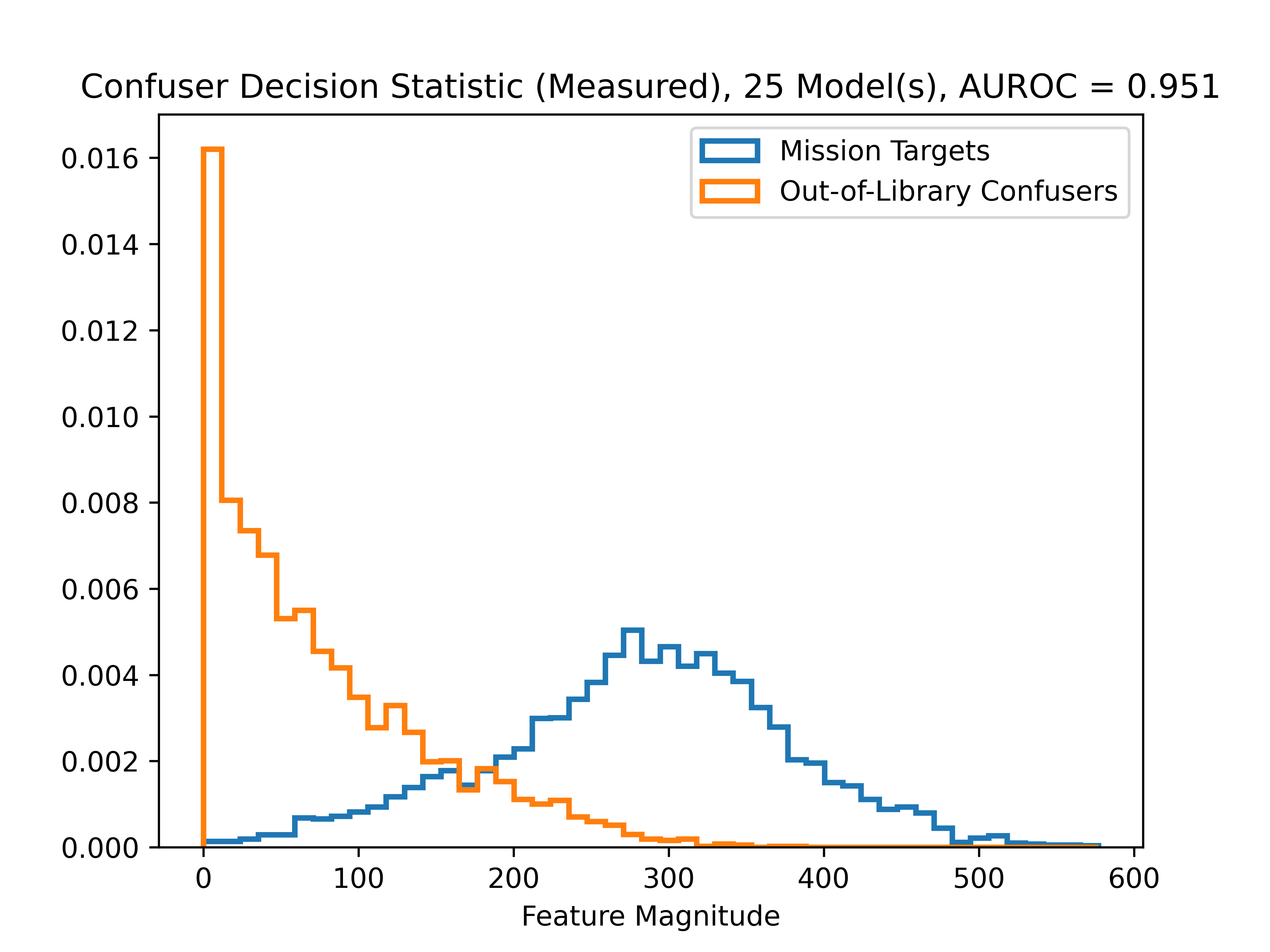}
        \caption{Weighted Model Calibration: $I=250$}
    \end{subfigure}
    \caption{Histograms comparing the feature magnitudes of mission targets to the magnitudes of out-of-library confusers for ensembled models.}
    \label{fig:fm_ensm}
\end{figure}

We now observe the feature magnitudes when we introduce both Unweighted Model Averaging and Weighted Model Calibration for $M=25$ and $I=0,250$ as shown in Figure~\ref{fig:fm_ensm}. While ensembling techniques for $I=0$ have minimal effect at separating the feature magnitudes between mission targets and confusers, models trained with $I=250$ show a much larger disparity in $F$ when compared to no ensembling. This shows that given some measured data and using ensembling, we can effectively isolate $F$ values of mission targets and confusers to generate more accurate results.

\subsubsection{Summary}

In this section, we made several observations in regards to performance when using different amounts of training measured data, different ensembling techniques, and different numbers of models ensembled. Table~\ref{tbl:overview} shows of the important values collected from the aforementioned experiments for easier review. We can see that in each key metric observed, ensembling methods out-performed a single model when some measured data is included in training compared to none. This is most likely due to the poor accuracy of the individual models, with ensembling only reinforcing the poor performance.

\begin{table}[h]
    \centering
    \renewcommand{\arraystretch}{1.2}
    \begin{tabular}{|c|c|c||c|c|c|}
        \hline
        \multirow{3}{*}{} & \multirow{3}{*}{$M$} & \multirow{3}{*}{$I$} & \multicolumn{3}{c|}{Ensembling Types} \\ 
        \cline{4-6}
        &  &  & Baseline & \parbox{3cm}{\centering Unweighted Model Averaging} & \parbox{3cm}{\centering Weighted Model Calibration} \\ 
        \hline
        \multirow{4}{*}{\parbox{3cm}{\centering AUROC}} & \multirow{2}{*}{1} & 0 & 0.823  & -  & -  \\ \cline{3-6}
                            &                     & 250 & 0.912  & -  & -  \\ \cline{2-6}
                            & \multirow{2}{*}{25} & 0 & -  & 0.797  & 0.813  \\ \cline{3-6}
                            &                     & 250 & -  & \textbf{0.952} & 0.951  \\  
        \hline \hline
        \multirow{4}{*}{\parbox{3cm}{\centering Mission Target Classification Accuracy}} & \multirow{2}{*}{1} & 0 & 0.4317 & - & - \\ \cline{3-6}
                            &                     & 250 & 0.8621 & - & - \\ \cline{2-6}
                            & \multirow{2}{*}{25} & 0 & -  & 0.3518 & 0.3251 \\ \cline{3-6}
                            &                     & 250 & -  & \textbf{0.9217} & 0.9180 \\
        \hline \hline
        \multirow{4}{*}{\parbox{3cm}{\centering Overall\\Classification Accuracy}} & \multirow{2}{*}{1} & 0 & 0.4123 & - & - \\ \cline{3-6}
                            &                     & 250 & 0.7096 & - & - \\ \cline{2-6}
                            & \multirow{2}{*}{25} & 0 & -  & 0.3163 & 0.3935 \\ \cline{3-6}
                            &                     & 250 & -  & 0.7399 & \textbf{0.7628} \\ 
        \hline
    \end{tabular}
    \caption{Results overview of accuracy and AUROC statistics}
    \label{tbl:overview}
\end{table}

\section{Conclusion}

In our paper, we attempted to solve two key challenges that many deep learning models face today: a lack of quality data and the rejection of out-of-distribution data. To supplement a lack of this data, we used a data generator to create synthetic data which has a similar distribution to the measured data trained on. Given an abundance of synthetically generated data, we can supplement our training with small amounts of measured data to drastically improve model performance even in small quantities.

To further bolster our model performance, we introduced two ensembling techniques: Unweighted Model Averaging and Weighted Model Calibration. Each considers the outputs from identically trained models, introducing diversity to our model outputs for both classification and confuser rejection purposes. We then observed that mission target classification and confuser rejection both saw excellent performance increases due to our tested methods.

Confuser rejection was also a core component of the paper, in which we wanted to analyze how our techniques performed at accurately rejecting these out-of-distribution targets. We saw that both using measured data when training and introducing ensembling helped in separating mission targets from confusers. Furthermore, we used the Objectosphere Loss to help exemplify feature magnitudes and separate the confidence intensity of the mission targets and confusers. As shown, training with the Objectosphere Loss and combining its efficacy with ensembling and supplemented measured data proved to be highly effective at isolating these feature magnitude distributions. We also saw a large increase in our AUROC score due to both ensembling and supplementing measured data.

Given the need in modern deep learning applications for high quality data and the need to reject out-of-distribution data, we can conclude that both techniques of supplementing synthetically generated data with real, measured data and the introduction of model ensembling are both highly effective solutions to these problems.

\typeout{}
\bibliographystyle{spiebib}
\bibliography{report}

\end{document}